\pdfoutput=1

\documentclass[11pt]{article}

\usepackage[]{ACL2023}

\usepackage{times}
\usepackage{latexsym}

\usepackage[T1]{fontenc}

\usepackage[utf8]{inputenc}

\usepackage{microtype}

\usepackage{inconsolata}

\usepackage{longtable}
\usepackage{tabu}
\usepackage{arydshln}

\usepackage{graphicx}
\usepackage{wrapfig}
\usepackage{CJK,algorithm,algorithmic,amssymb,amsmath,array,epsfig,graphics,float,subcaption,verbatim,epstopdf}
\usepackage{enumitem}
\usepackage{color,soul}
\usepackage{hhline}
\usepackage{multirow}
\usepackage{xcolor}
\usepackage{hyperref}
\usepackage{cleveref}

\usepackage{wrapfig,lipsum}

\usepackage{xurl}

\usepackage{tabularx}
\usepackage{bm}
\usepackage{seqsplit}
\usepackage{stmaryrd}
\usepackage{booktabs}

\usepackage{comment} 

\usepackage{xparse}
\title{Multimedia Generative Script Learning for Task Planning 
}

\author{
Qingyun Wang$^{1}$, \ Manling Li$^1$, \ Hou Pong Chan$^{2}$, \ Lifu Huang$^{3}$, \\ \  \textbf{Julia Hockenmaier}$^1$, \  \textbf{Girish Chowdhary}$^1$, \ \textbf{Heng Ji}$^{1}$\\   
$^{1}$ University of Illinois at Urbana-Champaign $^{2}$ University of Macau $^{3}$ Virginia Tech\\
$^{1}$\texttt{\fontfamily{pcr}\selectfont\{qingyun4,manling2,juliahmr,girishc,hengji\}@illinois.edu}\\
$^{2}${\fontfamily{pcr}\selectfont hpchan@um.edu.mo},$^{3}${\fontfamily{pcr}\selectfont lifuh@vt.edu}
}

\begin{document}
\maketitle
\begin{abstract}

Goal-oriented generative script learning aims to generate subsequent steps to reach a particular goal, which is an essential task to assist robots or humans in performing stereotypical activities. An important aspect of this process is the ability to capture historical states visually, which provides detailed information that is not covered by text and will guide subsequent steps. Therefore, we propose a new task, Multimedia Generative Script Learning, to generate subsequent steps by tracking historical states in both text and vision modalities, as well as presenting the first benchmark containing 5,652 tasks and 79,089 multimedia steps. This task is challenging in three aspects: the multimedia challenge of capturing the visual states in images, the induction challenge of performing unseen tasks, and the diversity challenge of covering different information in individual steps. We propose to encode visual state changes through a selective multimedia encoder to address the multimedia challenge, transfer knowledge from previously observed tasks using a retrieval-augmented decoder to overcome the induction challenge, and further present distinct information at each step by optimizing a diversity-oriented contrastive learning objective.  We define metrics to evaluate both generation and inductive quality. Experiment results demonstrate that our approach significantly outperforms strong baselines\footnote{The programs, data, and resources are publicly available for research purposes at: \url{https://github.com/EagleW/Multimedia-Generative-Script-Learning}.}. 
\end{abstract}


\section{Introduction}

\begin{figure}[htb!]
\centering
\includegraphics[width=0.9\linewidth]{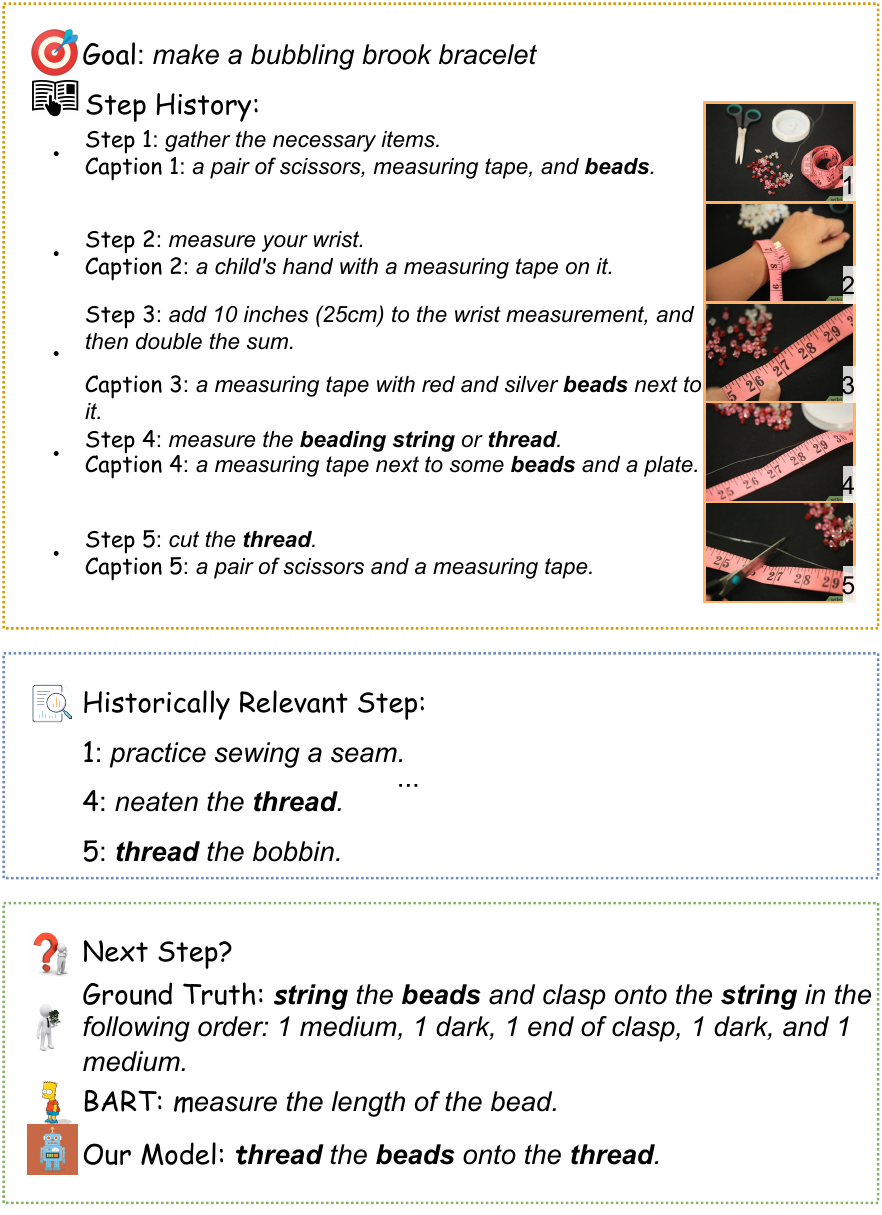}
\caption{\textbf{Multimedia Generative Script Learning:} The upper box shows the task input, including the goal and multimedia step history. Each step contains a text description and an illustrative image. The output is the next step. 
We retrieve historically relevant steps from the training corpus.  } 
\label{img:task_example}
\end{figure}

Robots rely on understanding the present real-world state and predicting the subsequent steps to better assist humans in daily stereotypical tasks such as meal preparation and gardening~\citep{10.1007/978-981-15-7345-3_30,9720489}. As an example, Robohow~\citep{robohow} uses articles from WikiHow\footnote{\url{https://www.wikihow.com} contains steps for a variety of tasks. } to assist robots in everyday tasks in human working and living environments. However, the problem is that not all daily tasks are well documented. Thus, generating a sequence of steps that lead to a given goal (i.e., goal-oriented generative script learning)~\citep{lyu-etal-2021-goal,huang2022language, zoey1, zoey2, zoey3} has a fundamental importance in allowing robots to perform unseen tasks by understanding the patterns in previously observed similar tasks. 

\begin{figure}[!bt]
\centering
\includegraphics[width=0.9\linewidth]{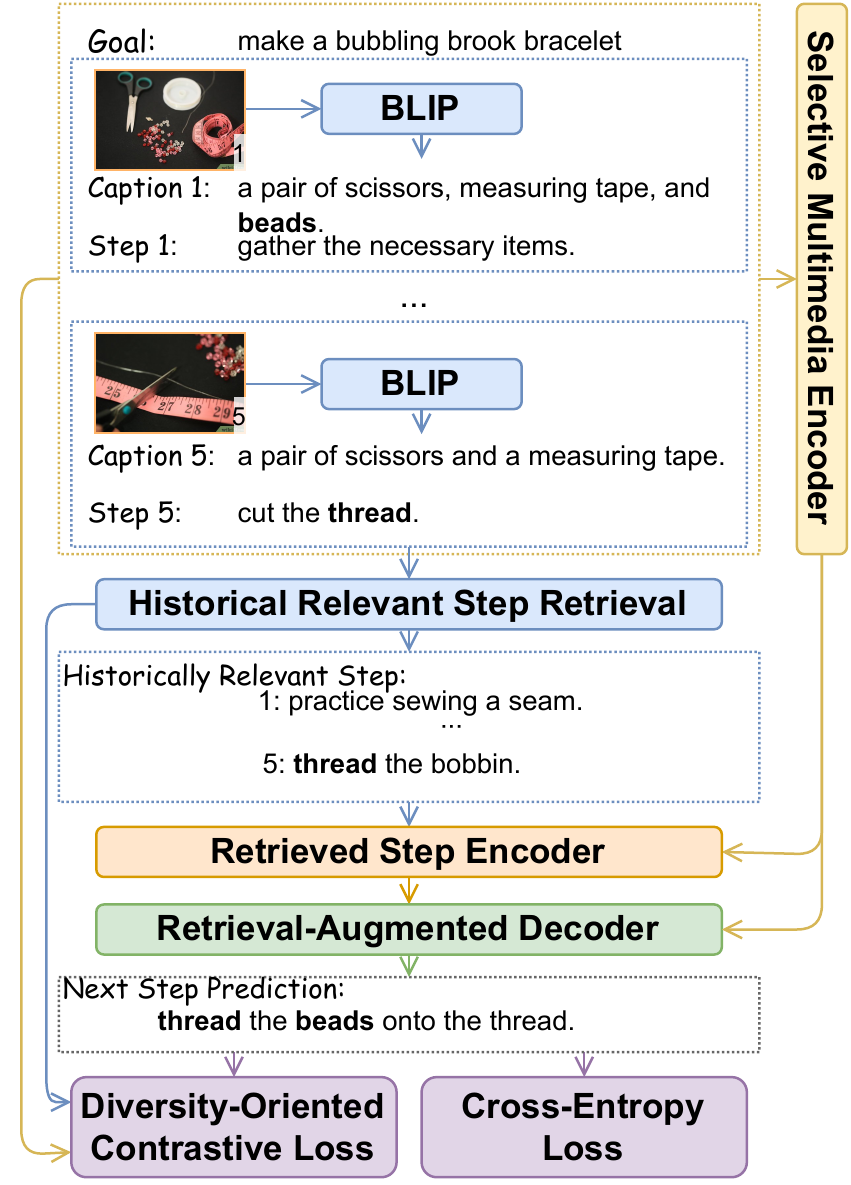}
\caption{Architecture overview. We use the example in Figure \ref{img:task_example} as the walking-through example. }
\label{img:overview}
\end{figure}

Despite this, previous goal-oriented generative script learning focuses solely on text~\citep{lyu-etal-2021-goal,huang2022language}, which is commonly affected by reporting bias~\citep{10.1145/2509558.2509563} as important details may be omitted in the source text. However, such information is often implicitly contained in images. For example, in Figure~\ref{img:task_example}, the image of Step 1 illustrates the items needed to \textit{make a bracelet}, which is not mentioned in the text but helps predict the action of \textit{threading beads} as a future step. 
Existing multimedia script learning work seeks to bridge this cross-media gap, but the task settings are multi-choice selection~\citep{yang-etal-2021-visual} or ordering~\citep{wu-etal-2022-understanding}, which require candidate steps as input so it is not a practical setting for real-life robots.

To address these problems, we propose a new task, \textbf{Multimedia Generative Script Learning} (Figure~\ref{img:task_example}), that requires systems to generate future steps based on the goal and previous steps with visual scenes depicting their states. Specifically, given the goal and previous step history in the form of natural language sentences paired with descriptive images, the model should automatically generate the natural language instruction for the next step.
A good script has three hallmarks:

(1) \underline{\textit{Visual-State Trackable}}: it records the historical visual scenes and recognizes significant changes that impact future steps. We call it \textit{multimedia challenge}.  To address this challenge, we focus on salient differences in visual scenes, and propose a novel \textbf{selective multimedia encoder}. Rather than learning directly from the visual details of each object, we first leverage an image captioner as an abstract summary of the image about global interactions among multiple objects. We then introduce a selection gate to focus on the selected captions and steps closely related to the future step. For instance, the second caption \textit{``a child's hand with a measuring tape on it''} in Figure ~\ref{img:task_example} can be filtered out by the selection gate because it is not closely related to the future steps.

(2) \underline{\textit{Inductive}}: it transfers knowledge from a previously observed task to similar unseen tasks. We call it \textit{induction challenge}. To induce procedural knowledge from previously observed tasks, we propose a \textbf{retrieval augmented decoder} to obtain relevant steps to guide the subsequent step generation. For example, the future step in Figure~\ref{img:task_example} closely resembles the scripts used in previous retrieved steps about \textit{threading items}, thus transferring script knowledge to an unseen task. 

(3) \underline{\textit{Diverse}}: it displays distinct information at each step.  We call it \textit{diversity challenge}. 
Existing pre-trained transformer-based language models such as T5 \citep{JMLR:v21:20-074}, BART \citep{lewis-etal-2020-bart}, and GPT-2 \citep{radford2019language} tend to generate repeated or highly similar future steps as shown in Figure~\ref{img:task_example}.
Therefore, we introduce a novel \textbf{diversity-oriented contrastive learning objective} to control all subsequent steps to convey different information. 
We treat all other steps in the given input and retrieved steps in other tasks similar to the given input as \textit{hard} negatives.

In addition to traditional generation-based metrics to evaluate task performance, we propose a new \textit{multimodal-retrieval based metric} to capture cross-modal semantic similarity. 
While the model design can be applied to any domain of interest, we experiment with the model on two domains \textit{Gardening} and \textit{Crafts}, where task planning has not been well researched. 
Automatic evaluation shows that our generated step predictions are close to the human written ground truth. Human evaluation further confirms that our diversity-oriented contrastive learning objective leads to diverse and correct steps.

The contributions are threefold: 
\begin{enumerate}
\item We propose the first \textit{multimedia goal-oriented generative script learning task} to record historical steps in both text and images. We also release a new benchmark from WikiHow, featuring 5,652 tasks and 79,089 multimedia steps.
\item We propose a novel approach to produce \textit{visually trackable}, \textit{inductive}, and \textit{diverse} scripts through a selective multimedia encoder, a retrieval augmented decoder, and a diversity-oriented contrastive learning objective. 
\item We propose a new \textit{multimodal-retrieval based metric} to evaluate the cross-modal semantic similarity and the inductive ability by checking factual correctness.
\end{enumerate}

\section{Problem Formulation}
We propose a new multimedia generative script learning task: given an activity goal $G$, an optional subgoal $M$ that specifies the concrete needs, and the previous multimedia step history $\mathcal{H}_n=\{(S_1,V_1),...,(S_n,V_n)\}$ with length $n$, a model is expected to predict the next possible step $S_{n+1}$, where $S_i$ is a text sequence and $V_i$ is an image.

\begin{table}[!htb]
\centering
\small
\begin{tabularx}{\linewidth}{>{\hsize=1.2\hsize}X>{\arraybackslash\hsize=0.9\hsize}X>{\centering\arraybackslash\hsize=0.8\hsize}X>{\centering\arraybackslash\hsize=0.8\hsize}X>{\centering\arraybackslash\hsize=1\hsize}X>{\centering\arraybackslash\hsize=1.2\hsize}X}
\toprule
\textbf{Domain}&\textbf{Split}  & \textbf{\#Task} & \textbf{\#Pair}& $\mathbf{\overline{\#Step}}$ & $\mathbf{\overline{\#Token}}$ \\
\midrule
                &Train           & 1,857 & 20,258 & 3.10 & 11.6 \\
Gardening       &Valid.      &  237  & 2,428  & 3.03 & 10.6\\
                &Test            &  238  & 2,684  & 2.88 & 11.2 \\
\hdashline
                &Train           &  2,654  & 32,082  & 6.06 & 8.98 \\
Crafts          &Valid.      &  3,33   & 4,061   & 6.12 & 9.10 \\
                &Test            &  3,33   & 3,937   & 5.91 & 9.00 \\
\bottomrule
\end{tabularx}
\caption{Statistics of our dataset. $\mathbf{\overline{\#Step}}$ denotes average number of steps per sample. $\mathbf{\overline{\#Token}}$ denotes average number of words per step.\label{tab:stat} }
\end{table}
\section{Dataset Collection}

Using articles from \textit{Gardening} and \textit{Crafts} categories as case studies, we create a new dataset based on the English WikiHow dump (2021/05). There are typically three levels of hierarchy in a WikiHow article: \textit{goals} which describe the overall task, \textit{subgoals} which represent the intermediate process to  accomplish a \textit{goal}, and \textit{steps} which are the specific actions to complete a \textit{subgoal}. For each WikiHow article, we collect step-image pairs as well as their goals and methods\footnote{We only keep steps that contain both images and texts.}. We split the whole dataset based on the task categories. Therefore, the validation and test sets contain tasks not included in the training set. Table \ref{tab:stat} shows the detailed data statistics.


\section{Method}

\subsection{Model Architecture}

The overall framework is illustrated in Figure \ref{img:overview}. 
Given the activity goal $G$, optional subgoal $M$, and multimedia step history $\mathcal{H}_n$, 
we first use an image captioner to map each input image into a precise caption and produce the caption-enhanced step history $\mathcal{\hat{H}}_{n}$. 
Then we propose a \textit{selective multimedia encoder} by extending the BART encoder with a gated fusion layer to learn contextualized representations for the step history. 
After that, a retrieval module retrieves historically relevant steps from the training corpus and encodes them with a \textit{retrieved step encoder}. 
Finally, we introduce a \textit{retrieval-augmented decoder}, which enhances the BART decoder with a retrieval gate fusion layer to fuse the representations of the input step history and retrieved steps to generate the next step.
The entire model is trained by our proposed \textit{diversity-oriented contrastive loss} and cross-entropy loss.

\subsection{Selective Multimedia Encoder}
\textbf{Image Encoding} Compared to step descriptions which focus more on action description, captions provide more visual environment/object information such as \textit{beads} in Step 1 from Figure \ref{img:overview}.
Because we are more concerned with the overall semantics of the salient objects in the image rather than the details of every object, we adopt image captioners to encode visual features and track visual state changes. For instance, while multiple objects are present in Step 3 in Figure \ref{img:task_example}, the \textit{finger} object can be ignored in the third step as it does not represent the key information conveyed by the image. Specifically, we use the state-of-the-art image captioner BLIP~\citep{li2022blip}, which is pretrained on a large-scale vision-and-language corpus with 129M images to generate a caption $C_{i}$ for each image $V_{i}$ in the input step history $\mathcal{H}_{n}$. After that, we obtain the \textit{caption-enhanced step history} $\mathcal{\hat{H}}_{n}=\{(S_1,C_1),...,(S_n,C_n)\}$, where $C_i$ is the caption of the image $V_i$ in step $i$.

\textbf{Selective Multimedia Encoding}
To help the encoder capture the activity goal and subgoal information, we concatenate goal $G$ and optional subgoal $M$ to serve as the first sequence in the history $X_0= [G, M]$. For the subsequent steps in the history, we concatenate each step and caption as $X_{2i-1}=S_i$ and $X_{2i}=C_i$. To summarize the step history, we prepend a learnable $\mathtt{[CLS]}$ token to the sequence as a contextualized vector. The entire text sequence is then represented as $\mathcal{X}=\{\mathtt{[CLS]},X_0,X_1,...,X_{2n}\}$. We pass the text sequence $\mathcal{X}$ into a BART encoder to get the contextualized hidden representation $\mathbf{H}=\{\mathbf{h}_0,...,\mathbf{h}^{2n}_{L_{X_{2n}}}\}=\mathrm{Enc}(\mathcal{X})$. We denote $\mathbf{H}_{X_j}=\{\mathbf{h}^j_1,...,\mathbf{h}^j_{L_{X_j}}\}$ as the hidden states for sequence $X_j$, where $L_{X_j}$ is the length of  $X_j$.

Since the input sequence contains steps or captions not directly relevant to the future step, we need to mask those sentences based on the step/caption representations. For instance, in Figure \ref{img:overview}, the step description for Step 1 is vague and needs to be masked. We treat the representation of the $\mathtt{[CLS]}$ token, $\mathbf{h}_0$, as the contextualized representation of the entire step history and use it to compute a mask that filters out the irrelevant step/caption information.  Specifically, we use $\mathbf{h}_0$ as query and $\mathbf{H}_{X_j}$ as both the key and value to compute Multi-Headed Attention ($\mathrm{MultiHead}$) \citep{NIPS2017_3f5ee243} for each sequence hidden states $\mathbf{H}_{X_j}$: $\hat{\mathbf{h}}_{X_j} = \mathrm{MultiHead}(\mathbf{h}_0,\mathbf{H}_{X_j},\mathbf{H}_{X_j})$,
where $\mathbf{\hat{h}}_{X_j}$ is the weighted representation for text sequence $X_j$.
Then, for each sequence $X_j$, we can calculate the mask probability as:$
\alpha_j=\sigma (\mathbf{W}_\alpha[\mathbf{\mathbf{h}_0;\hat{h}}_{X_j}])$, where $\mathbf{W}_\alpha$ is a learnable parameter. Similar to \citet{sengupta-etal-2021-gated-transformer}, 
we update the hidden states for each sequence $X_j$ as $
\mathbf{\bar{H}}_{X_j} = \alpha_j \cdot \mathbf{emb}_{\mathtt{[MASK]}} + (1-\alpha_j )\mathbf{H}_{X_j}
$,
where $\mathbf{emb}_{\mathtt{[MASK]}}$ is the embedding of the $\mathtt{[MASK]}$ token. 
The final hidden state sequences are  $\mathbf{\bar{H}}=[h_0;\mathbf{\bar{H}}_1;...;\mathbf{\bar{H}}_{2n}]$.

\subsection{Step Retrieval Augmentation} 
\label{sec:retrieve}
\textbf{Historically Relevant Step Retrieval} In addition to the caption-enhanced step history, $\mathcal{\hat{H}}_{n}$, we retrieve historically relevant steps $\mathcal{R}_{n+1}= \{R_1, ..., R_k\}$ from the training tasks, where $k$ is the number of retrieved relevant steps. We first use SentenceBERT~\citep{reimers-gurevych-2019-sentence} to encode all steps. We then retrieve $k$ steps from the training corpus, which have the top-k highest cosine similarity to the previous step $S_n$ from the representation given by SentenceBERT\footnote{We use the previous step $S_n$ instead of all history since it is more temporally correlated to the next step.}. Finally, we consider the immediate next step for each of those $k$ steps as potential relevant steps $\mathcal{R}_{n+1}$. For instance, because Step 5 in Figure~\ref{img:overview} is similar to \textit{pull the thread out} in the training corpus, we choose its immediate next step \textit{thread the bobbin} as a historically relevant step.

\noindent\textbf{Retrieved Step Encoder} For historically relevant steps $\mathcal{R}= \{R_1, ..., R_k\}$, we apply the BART encoder to get hidden states $\mathbf{H}_R=\{\mathbf{H}_{R_1};....;\mathbf{H}_{R_k}\}$. 
Similarly, we use $\mathbf{h}_0$ in multimedia encoder as the query and $\mathbf{H}_{R_i}$ as both the key and value to compute multi-headed attention for each sequence hidden states: $\hat{\mathbf{h}}_{R_i}=\mathrm{MultiHead}(\mathbf{h}_0,\mathbf{H}_{R_i},\mathbf{H}_{R_i})$,
where $\mathbf{\hat{h}}_{R_i}$ is the weighted representation for step sequence $R_i$. 
Similarly, we can calculate the mask probability as: $\beta_j=\sigma (\mathbf{W}_\beta[\mathbf{h}_0;\mathbf{\hat{h}}_{R_j}])$, 
where $\mathbf{W}_\beta $ is a learnable parameter. We then update the hidden states for each sequence $R_j$ as $\mathbf{\bar{H}}_{R_i} = \beta_j \cdot \mathbf{emb}_{\mathtt{[MASK]}} + (1-\beta_j )\mathbf{H}_{R_i}$.
The final hidden state sequences is $\mathbf{\bar{H}}_R=[\mathbf{\bar{H}}_{R_1};...;\mathbf{\bar{H}}_{R_k}]$.

\subsection{Retrieval-Augmented Decoder}

In the decoder, we compute the probability $P\left(s_{q}|s_{<q},\mathcal{\hat{H}},G,M\right)$ for the $q$-th token $s_q\in S_{n+1}$.
Our retrieval-augmented decoder is similar to \cite{liu-etal-2021-three}, which aims to capture historically relevant steps related to the next step based on previous decoder hidden states. Given $z_q^l$ which is the hidden state of $s_{q}$ in layer $l$, we first use a multi-head cross-attention to fuse the hidden states from the retrieved steps $\mathbf{\bar{H}_R}$: $  {z'_q}^l = \mathrm{MultiHead}(z_q^l,\mathbf{\bar{H}}_R,\mathbf{\bar{H}}_R)$.
We also append a gating mechanism to control the knowledge from the retrieved steps and previous hidden states:
\begin{equation}
\begin{split}
    \gamma &= \sigma(\mathbf{W}_\gamma[z_q^l;{z'_q}^l])    \\
    {\tilde{z}_q}^l &= \gamma \cdot \mathrm{LN}({z'_q}^l) + (1-\gamma)\cdot (z_q^l)
\end{split}
\end{equation}
where $\mathbf{W}_\gamma$ is a learnable parameter and $\mathrm{LN}(*)$ is the layer norm function. 
Finally, the fused hidden states in the top layer are used to compute the generation probability.
We supervise the next step generation using the standard cross-entropy loss: 
\begin{equation}
\mathcal{L}_{\mathrm{gen}} = \sum_{q=1}^{|S_{n+1}|} \log P\left(s_{q}|s_{<q},\mathcal{\hat{H}},G,M\right)
\end{equation}

\subsection{Diversity-Oriented Contrastive Learning}

In the experiment, we observe that the model tends to keep generating similar future steps in a row given the beginning steps as input or just paraphrases the input steps. 
Therefore, we propose a contrastive learning-based loss to encourage the model to return diverse step prediction results.

\noindent\textbf{Negative Sampling} Sequence-to-sequence models suffer from the ``exposure bias'' problem \citep{ranzato2015sequence,an2022cont} because of \textit{teacher forcing}. Contrastive loss provides an additional sequence level loss which can help models increase the diversity of the output steps. We adopt two types of negative sampling strategies to discourage the model from paraphrasing the previous step as the future step: 
\textit{self-negatives} \citep{wang-etal-2022-simkgc} where we consider the input steps as negative samples and \textit{retrieved-negatives} where we consider the retrieved steps from training corpus which are similar to the input step as negative samples. 
For example, in Figure \ref{img:task_example}, the goals and steps from the step history serve as the self-negatives. Given the last step, “cut the thread”, we retrieve similar steps from the training set as retrieved negatives which include “cut your thread”, "cut off the extra thread", etc.

\noindent\textbf{Diversity-Oriented Contrastive Loss} Since the model needs to distinguish between the ground truth and those negative samples, we design a novel diversity-oriented contrastive loss. Specifically, given an input sequence $\mathcal{\hat{H}},G,M$, the ground truth next step $S_{n+1}$, and a set of $K$ negative samples $\{S_{n+1}^1, S_{n+1}^2,...,S_{n+1}^K\}$, we aim to maximize the probability of classifying the positive sample correctly with the InfoNCE loss \citep{oord2018representation}:
\begin{align}
\begin{split}
    \mathcal{L}_{\mathrm{cl}} &= \frac{\exp{\left(y^+/\tau\right)}}{\sum_k \exp{\left(y^-_k/\tau\right)} +\exp{\left(y^+/\tau\right)} } \\
    y^+&=\sigma(\mathrm{Avg}(\mathbf{W}_y\mathbf{\bar{H}}^++\mathbf{b}_y))\\
    y^-_k&=\sigma(\mathrm{Avg}(\mathbf{W}_y\mathbf{\bar{H}}^-_k+\mathbf{b}_y))\\
\end{split}
\end{align}
where $\mathbf{\bar{H}}^+$ and $\mathbf{\bar{H}}_k^-$ are decoder hidden states from the positive and $k$-th negative samples, $\mathbf{W}_y$ is a learnable parameter, $\tau$ is the temperature, and $\mathrm{Avg}(*)$ denotes the average pooling function.

\subsection{Training Objective}

We jointly optimize the cross-entropy loss and our proposed diversity-oriented contrastive loss: $\mathcal{L} = \mathcal{L}_{\mathrm{gen}} + \lambda \mathcal{L}_{\mathrm{cl}}$, 
where $\lambda$ is a hyperparameter that controls the weight of the contrastive loss.

\section{Evaluation Metrics}

\textbf{Generation Quality Evaluation} Following common practice in text generation, we first evaluate our model with BLEU \citep{papineni-etal-2002-bleu}, ROUGE \citep{lin-2004-rouge}, and METEOR \citep{denkowski-lavie-2014-meteor} scores to examine the content overlap between generated steps and ground truth. 

\noindent
\textbf{Inductive Quality Evaluation} 
In order to determine whether the inferred subsequent steps are factually correct, we further evaluate the models with BARTScore \citep{NEURIPS2021_e4d2b6e6} and the semantic similarity score \citep{thakur-etal-2021-augmented}. The semantic similarity score uses a cross-encoder pretrained on STSBenchmark \citep{cer-etal-2017-semeval} to calculate the semantic similarity between two sentences.

In addition to evaluating whether the generated step matches the next step, we also check whether the generated step matches any subsequent step. 
This enables the model to earn credit if it generates a step that appears in the future.
We propose a \textit{Multimodal-Retrieval based metric}: for each generated step, we use it as a query to search all corresponding step-image pairs under the same subgoal/goal from the testing set. We then compute HIT@1 for results that fall into ground-truth future step-image pairs. Similar to Section \ref{sec:retrieve}, we use SBERT~\citep{reimers-gurevych-2019-sentence} to rank the most similar steps under the same subgoal to get Text@1 (T@1). To compute Image@1 (I@1), we use CLIP \citep{pmlr-v139-radford21a} to rank the most similar images under the same subgoal. If the top-1 retrieval results appear in the subsequent steps, we consider it a HIT. The retrieval-based metric captures normalized semantic similarity concerning all related steps under certain subgoals. The CLIP-based retrieval metric also enables the evaluation of the cross-modality semantic similarity. 
Additional details of the evaluation setup are in the Appendix \ref{sec:evalmetrics}.

\begin{table}[!htb]

\centering
\small
\begin{tabularx}{\linewidth}{>{\hsize=2.6\hsize}X>{\centering\arraybackslash\hsize=0.6\hsize}X>{\centering\arraybackslash\hsize=0.6\hsize}X>{\centering\arraybackslash\hsize=0.6\hsize}X>{\centering\arraybackslash\hsize=0.6\hsize}X}
\toprule
\multirow{ 2}{*}{\textbf{Model}}&\multicolumn{2}{c}{\textbf{Gardening}}&\multicolumn{2}{c}{\textbf{Crafts}}\\
                                &\textbf{I@1}$\uparrow$&\textbf{T@1}$\uparrow$&\textbf{I@1}$\uparrow$&\textbf{T@1}$\uparrow$\\ 
\midrule
BART                           &  44.6              & 40.0          &48.2           & 29.9 \\
+CP               &  48.5              & 39.2          &48.2           & 31.5\\ 
+CP+M            &  \textbf{49.8}     & 41.0          &\textbf{50.3}  & \textbf{37.8}\\ 
+CP+M+R         & 48.1               & 38.9          &48.9           & 31.8\\ 
+CP+M+R+CL      &  49.5              & \textbf{43.0} &49.0           & 33.9\\
\bottomrule
\end{tabularx}
\caption{Multimodal-retrieval based evaluation (\%). \textit{CP} is models with caption input. \textit{M} is models with selective multimedia encoder. \textit{R} is models with historically relevant step encoder and retrieval-augment decoder. \textit{CL} is models with diversity-oriented contrastive learning.
}
\label{tab:fut}
\end{table}

\begin{table*}[!htb]
\centering
\small

\begin{tabularx}{\linewidth}{>{\hsize=2.4\hsize}X>{\centering\arraybackslash\hsize=0.6\hsize}X>{\centering\arraybackslash\hsize=0.6\hsize}X>{\centering\arraybackslash\hsize=0.6\hsize}X>{\centering\arraybackslash\hsize=0.6\hsize}X>{\centering\arraybackslash\hsize=1.1\hsize}X>{\centering\arraybackslash\hsize=0.7\hsize}X>{\centering\arraybackslash\hsize=1.2\hsize}X>{\centering\arraybackslash\hsize=1.2\hsize}X}
\toprule
\textbf{Model}&\textbf{B-1}$\uparrow$&\textbf{B-2}$\uparrow$&\textbf{B-3}$\uparrow$&\textbf{B-4}$\uparrow$&\textbf{METEOR}$\uparrow$&\textbf{R-L}$\uparrow$&\textbf{BARTScore}$\uparrow$&\textbf{Semantic}$\uparrow$\\  
\midrule
GPT-2                       & 13.2 & 5.03 & 1.87 & 0.72 & 7.38 & 12.5 & -4.73 & 0.239\\
T5                          & 17.6 & 9.05 & 4.92 & 2.87 & 9.41 & 16.5 & -4.45 & 0.300\\
Naive Retrieval             & 10.9 & 4.14 & 1.93 & 1.10 & 6.33 & 10.0 & -4.88 & 0.180\\
CLIP-BART                   & 14.4 & 7.10 & 3.77 & 2.22 & 8.28 & 13.8 & -4.44 & 0.256\\
Retrieval BART    & 16.8 & 8.68 & 4.80 & 2.24 & 9.15 & 16.0 & -4.43 & 0.295\\
GPT2-SIF          & 11.6 & 5.10 & 2.43 & 1.28 & 6.85 & 10.8 & -4.80 & 0.233\\
BART                        & 17.0 & 8.21 & 4.45 & 2.61 & 8.93 & 15.7 & -4.52 & 0.277\\
\hdashline
\hspace{1mm}+CP            & 16.9 & 8.79 & 4.99 & 3.03 & 9.23 & 16.5 & -4.41 & 0.300\\
\hspace{1mm}+CP+M        & 17.8 & 9.36 & 5.30 & 3.19 & 9.61 & \textbf{17.4} & -4.38 & 0.305\\
\hspace{1mm}+CP+M+R     & 17.5 & 9.22 & 5.25 & 3.13 & 9.60 & 17.2 & \textbf{-4.36} & 0.309\\
\hspace{1mm}+CP+M+R+CL  & \textbf{18.4} & \textbf{9.72} & \textbf{5.51} & \textbf{3.31} & \textbf{9.91} & 17.3 & -4.37 & \textbf{0.310}\\
\bottomrule
\end{tabularx}
\caption{Results with automatic evaluation on next step prediction for the gardening domain (\%). \textit{B-n} denotes the BLEU-n score. \textit{R-L} denotes the ROUGE-L score. \textit{Semantic} denotes semantic similarity score.
\label{tab:step} }
\end{table*}

\begin{table*}[!htb]
\centering
\small

\begin{tabularx}{\linewidth}{>{\hsize=2.4\hsize}X>{\centering\arraybackslash\hsize=0.6\hsize}X>{\centering\arraybackslash\hsize=0.6\hsize}X>{\centering\arraybackslash\hsize=0.6\hsize}X>{\centering\arraybackslash\hsize=0.6\hsize}X>{\centering\arraybackslash\hsize=1.1\hsize}X>{\centering\arraybackslash\hsize=0.7\hsize}X>{\centering\arraybackslash\hsize=1.2\hsize}X>{\centering\arraybackslash\hsize=1.2\hsize}X}
\toprule
\textbf{Model}&\textbf{B-1}$\uparrow$&\textbf{B-2}$\uparrow$&\textbf{B-3}$\uparrow$&\textbf{B-4}$\uparrow$&\textbf{METEOR}$\uparrow$&\textbf{R-L}$\uparrow$&\textbf{BARTScore}$\uparrow$&\textbf{Semantic}$\uparrow$\\ 
\midrule
GPT-2                       & 15.5 & 5.40 & 1.98 & 0.93 & 7.63  & 14.0 & -4.67 & 0.218\\
T5                          & 20.8 & 11.1 & 6.43 & 4.07 & 10.54 & 19.6 & -4.38 & 0.300\\
Naive Retrieval             & 13.5 & 5.26 & 2.38 & 1.28 & 6.81  & 12.3 & -4.83 & 0.163\\
CLIP-BART                   & 17.9 & 9.13 & 5.21 & 3.40 & 9.37  & 16.4 & -4.56 & 0.245\\
Retrieval BART              & 18.7 & 9.78 & 5.52 & 3.52 & 9.89  & 18.2 & -4.38 & 0.285\\
GPT2-SIF                    & 14.8 & 6.70 & 3.05 & 1.58 & 7.74  & 13.2 & -4.69 & 0.234\\
BART                        & 19.7 & 10.8 & 6.22 & 4.11 & 10.44 & 20.0 & -4.29 & 0.299\\
\hdashline
\hspace{1mm}+CP            & 20.1 & 11.1 & 6.48 & 4.24 & 10.61 & 20.1 & -4.29 & 0.303\\
\hspace{1mm}+CP+M         & 20.5 & 11.1 & 6.61 & 4.40 & 10.79 & 20.1 & -4.28 & 0.305\\
\hspace{1mm}+CP+M+R      & 20.7 & 11.5 & 6.93 & 4.66 & 11.02 & \textbf{20.5} & \textbf{-4.25} & 0.309\\
\hspace{1mm}+CP+M+R+CL   & \textbf{21.3} & \textbf{11.8} & \textbf{7.12} & \textbf{4.85} & \textbf{11.25} & 20.3 & -4.26 & \textbf{0.313}\\ 
\bottomrule
\end{tabularx}
\caption{Automatic evaluation results on next step prediction for the crafts domain (\%).
\label{tab:step1} }
\end{table*}

\section{Experiments}

\subsection{Baselines}

We first compare our model with \textbf{(1) state-of-the-art pretrained text-only generation models} to examine the results without tracking visual states, including GPT-2 \citep{radford2019language}, T5 \citep{JMLR:v21:20-074}, and BART \citep{lewis-etal-2020-bart}. We then compare our model with the \textbf{(2) retrieval baselines} including a naive retrieval baseline which directly uses retrieved historically relevant sentences as discussed in Section \ref{sec:retrieve}, and retrieval BART which takes in the concatenation of the retrieved historically relevant sentences with the original text input. We also include \textbf{(3) multi-modal generation baselines} that can take image embedding instead of captions as input, which is equivalent to CLIP-BART \citep{Sung_2022_CVPR}. The CLIP-BART has a similar backbone as VL-BART \citep{pmlr-v139-cho21a} but instead replacing the Faster R-CNN \citep{NIPS2015_14bfa6bb} with ViT-B/32 CLIP encoder \citep{pmlr-v139-radford21a} which has a better image-text alignment. Additionally, we compare our model with a state-of-the-art script learning model: GPT2-SIF \cite{sancheti-rudinger-2022-large} finetuned on our dataset. Finally, we include the variances of our model as \textbf{(4) baselines for ablation}. 
We select BART over T5 as the base model due to the performance and parameter size. Due to the large number of parameters in T5 (222M) compared to BART (139M), given similar model performance in Table \ref{tab:step} and \ref{tab:step1}, we choose BART instead of T5. 
The hyperparameters, training details, and additional ablation study are presented in the Appendix \ref{sec:hyper}, \ref{sec:train}, and \ref{sec:abl}.

\begin{table}[!htb]
\centering
\small
\begin{tabularx}{\linewidth}{>{\hsize=4.2\hsize}X>{\centering\arraybackslash\hsize=0.6\hsize}X>{\centering\arraybackslash\hsize=0.6\hsize}X>{\centering\arraybackslash\hsize=0.6\hsize}X>{\centering\arraybackslash\hsize=0.6\hsize}X>{\centering\arraybackslash\hsize=0.6\hsize}X>{\centering\arraybackslash\hsize=0.6\hsize}X>{\centering\arraybackslash\hsize=0.6\hsize}X>{\centering\arraybackslash\hsize=0.6\hsize}X}
\toprule
\multirow{ 2}{*}{\textbf{Model}}&\multicolumn{4}{c}{\textbf{Gardening}}&\multicolumn{4}{c}{\textbf{Crafts}}\\
&\textbf{1}$\downarrow$&\textbf{2}$\downarrow$&\textbf{3}$\downarrow$&\textbf{4}$\downarrow$&\textbf{1}$\downarrow$&\textbf{2}$\downarrow$&\textbf{3}$\downarrow$&\textbf{4}$\downarrow$\\
\midrule
Ground Truth                   & 37.0           & 3.08          & 0.42          & 0.18          & 30.6 & 1.07 & 0.05 & 0.00\\
\hdashline
BART                           & 45.2           & 6.94          & 1.39          & 0.73          & 39.2 & 2.18 & 0.26 & 0.10\\
+CP               & \textbf{43.1}  & 5.88          & 1.00          & 0.39          & \textbf{36.0} & \textbf{1.81} & 0.05 & 0.02 \\
+CP+M            & 43.6           & \textbf{5.75} & \textbf{0.78} & \textbf{0.20} & 36.4 & 1.97 & \textbf{0.02} & 0.01 \\
+CP+M+R         & 44.2           & 6.32          & 1.12          & 0.38          & 36.9 & 2.03 & 0.06 & \textbf{0.01} \\
+CP+M+R+CL      & 43.3           & 6.23          & 1.01          & 0.35          & 36.2 & 1.91 & 0.05 & 0.02 \\
\bottomrule
\end{tabularx}
\caption{Percent (\%) of $n$-grams in step history which appear in  human or system steps\label{tab:over}. }
\end{table}

\subsection{Automatic Evaluation}
As shown in Table \ref{tab:step} and \ref{tab:step1}, our model outperforms baselines. Since our task is open-ended and we are testing on unseen activities, our generated sentences usually contain paraphrases. Therefore, the BLEU scores, which rely on the exact word $n$-grams match \citep{inlg2018bleu}, are not high. In particular, because our ground truth only has an average length of 11 which contains less $4$-grams than the text in other tasks, our BLEU-4 is lower than other text generation tasks. The substantial gap between CLIP-BART and BART or BART with caption indicates that captions usually carry more specific information than images, and the current multimodal encoders still cannot perfectly embed text and images into the same semantic space. Meanwhile, the low performance of the retrieval baselines shows that simple retrieval methods are insufficient to predict accurate next steps. 

\begin{table}[!htb]
\centering
\small

\begin{tabularx}{\linewidth}{>{\hsize=4.2\hsize}X>{\centering\arraybackslash\hsize=0.6\hsize}X>{\centering\arraybackslash\hsize=0.6\hsize}X>{\centering\arraybackslash\hsize=0.6\hsize}X>{\centering\arraybackslash\hsize=0.6\hsize}X>{\centering\arraybackslash\hsize=0.6\hsize}X>{\centering\arraybackslash\hsize=0.6\hsize}X>{\centering\arraybackslash\hsize=0.6\hsize}X>{\centering\arraybackslash\hsize=0.6\hsize}X}
\toprule
\multirow{ 2}{*}{\textbf{Model}}&\multicolumn{4}{c}{\textbf{Gardening}}&\multicolumn{4}{c}{\textbf{Crafts}}\\
&\textbf{1}$\downarrow$&\textbf{2}$\downarrow$&\textbf{3}$\downarrow$&\textbf{4}$\downarrow$&\textbf{1}$\downarrow$&\textbf{2}$\downarrow$&\textbf{3}$\downarrow$&\textbf{4}$\downarrow$\\
\midrule
Ground Truth              & 87.1 & 60.1 & 36.1 & 23.6 & 91.3 & 68.7 & 41.6 & 27.7 \\
\hdashline
BART                      & 93.7 & 84.3 & 72.9 & 64.2 & 96.9 & 90.6 & 80.6 & 73.5 \\
+CP          & 92.8 & 81.3 & 68.9 & 60.5 & 96.3 & 89.3 & 79.2 & 72.5 \\
+CP+M       & 96.2 & 89.9 & 81.4 & 73.9 & \textbf{95.9} & \textbf{87.8} & 76.6 & 68.5 \\
+CP+M+R    & \textbf{92.3} & \textbf{80.5} & \textbf{67.9} & \textbf{57.8}& 96.9 & 89.6 & 78.6 & 71.1 \\
+CP+M+R+CL & 95.1 & 87.2 & 77.1 & 68.6 & 96.3 & 88.0 & \textbf{75.8 }& \textbf{67.3} \\
\bottomrule
\end{tabularx}
\caption{Self-BLEU (\%) for human or system steps\label{tab:selfbleu}. }
\end{table}

Among our model variants, adding selective encoding leads to a further performance increase, showing that selective encoding helps the model focus on the content in step history that is most related to future steps. The superior performance on BARTScore and semantic similarity of the retrieval-augmented model indicates the effectiveness of the guidance from historically relevant steps. Our contrastive learning model achieves larger gains compared to baselines for BLEU and METEOR, suggesting that our contrastive loss helps the model generate results similar to the ground truth. 

\noindent\textbf{Automatic Evaluation with Future Steps} 
We evaluate whether the predicted step is related to any future steps. Our contrastive learning model outperforms other ablations significantly on text retrieval for the Gardening domain, as shown in Table~\ref{tab:fut}. These results imply that the contrastive learning objective encourages the model to generate more informative future steps. The decrease in n-gram overlap between input step history and step predictions~(Table \ref{tab:over}) suggests that the contrastive learning objective also decreases the model's paraphrasing tendency. Interestingly, the performance decreases when adding the retrieval augmentation to the model because the retrieval model introduces additional information related to the step history, which makes the model generate results similar to previous steps~(Table \ref{tab:over}).

\begin{table}[!htb]
\centering
\small

\begin{tabularx}{\linewidth}{>{\hsize=4.2\hsize}X>{\centering\arraybackslash\hsize=0.6\hsize}X>{\centering\arraybackslash\hsize=0.6\hsize}X>{\centering\arraybackslash\hsize=0.6\hsize}X>{\centering\arraybackslash\hsize=0.6\hsize}X>{\centering\arraybackslash\hsize=0.6\hsize}X>{\centering\arraybackslash\hsize=0.6\hsize}X>{\centering\arraybackslash\hsize=0.6\hsize}X>{\centering\arraybackslash\hsize=0.6\hsize}X}
\toprule
\multirow{ 2}{*}{\textbf{Model}}&\multicolumn{4}{c}{\textbf{Gardening}}&\multicolumn{4}{c}{\textbf{Crafts}}\\
&\textbf{1}$\uparrow$&\textbf{2}$\uparrow$&\textbf{3}$\uparrow$&\textbf{4}$\uparrow$&\textbf{1}$\uparrow$&\textbf{2}$\uparrow$&\textbf{3}$\uparrow$&\textbf{4}$\uparrow$\\
\midrule
Ground Truth                   & 11.4           & 50.9          & 80.8          & 92.2& 8.46 & 44.4 & 77.9 & 90.9\\
\hdashline
BART                           & 4.75           & 17.7          & 32.4          & 42.6& 5.11 & 22.6 & 42.8 & 53.8  \\
+CP               & \textbf{5.17}  & 19.2          & 33.7          & 42.7& 5.12 & 22.6 & 42.7 & 53.8\\
+CP+M            &  4.94          & 18.6          & 32.8          & 41.8  & 4.92 & 22.4 & 42.3 & 53.8\\
+CP+M+R         &  5.06          & 19.2          & 34.6          & 44.3 & \textbf{5.23} & \textbf{23.3} & 43.9 & 55.2\\
+CP+M+R+CL      &  5.02          & \textbf{19.3} & \textbf{35.0} & \textbf{45.2} & 5.07 & 23.3 & \textbf{44.2} & \textbf{56.1}\\
\bottomrule
\end{tabularx}
\caption{Unique $n$-grams in human or system steps(\%). \label{tab:gram} }
\end{table}
\noindent\textbf{Automatic Evaluation on Diversity}  To evaluate the diversity between generated steps in the test sets, we employ two diversity metrics: self-BLEU \citep{10.1145/3209978.3210080}  (Table~\ref{tab:selfbleu}) and unique $n$-grams \citep{fedus2018maskgan} (Table~\ref{tab:gram}). The self-BLEU evaluates whether a model produces similar $n$-grams in different samples by measuring the similarity between one sentence and the rest in the test set. The retrieval model achieves the best results for the Gardening domain because it acquires additional knowledge from the retrieved steps and thus diversifies the output. The contrastive learning model achieves the best self-BLEU for 3,4 grams for the Crafts domain, implying our model's effectiveness. The unique $n$-grams calculate the percentage of distinct $n$-grams. It considers the repetition of n-grams within a generated step and across samples. The contrastive learning model achieves the highest distinct scores for 3,4 grams for both domains, indicating the effectiveness of our diversity-based contrastive loss in generating more diverse steps. 

\subsection{Human Evaluation}
\label{sec:human}

\begin{table}[!htb]
\centering
\small
\begin{tabularx}{\linewidth}{>{\hsize=4.2\hsize}X>{\centering\arraybackslash\hsize=0.6\hsize}X>{\centering\arraybackslash\hsize=0.6\hsize}X>{\centering\arraybackslash\hsize=0.6\hsize}X>{\centering\arraybackslash\hsize=0.6\hsize}X>{\centering\arraybackslash\hsize=0.6\hsize}X>{\centering\arraybackslash\hsize=0.6\hsize}X>{\centering\arraybackslash\hsize=0.6\hsize}X>{\centering\arraybackslash\hsize=0.6\hsize}X}
\toprule
\multirow{ 2}{*}{\textbf{Model}}&\multicolumn{4}{c}{\textbf{Gardening}}&\multicolumn{4}{c}{\textbf{Crafts}}\\
&\textbf{N.}$\downarrow$&\textbf{F.}$\downarrow$&\textbf{D.}$\downarrow$&\textbf{E.}$\downarrow$&\textbf{N.}$\downarrow$&\textbf{F.}$\downarrow$&\textbf{D.}$\downarrow$&\textbf{E.}$\downarrow$\\
\midrule
BART            & 1.92           & 2.05          &  2.43         & 1.60         & 1.90 & 2.03 & 2.29 & 1.76\\
+CP             & 1.78           & 1.93          &  2.70         & 1.39         & 1.70 & 1.85 & 2.86 & 1.65 \\
+CP+M          & 1.77           & 1.95          &  2.41         & 1.37         & 2.15 & 2.04 & 4.11 & 1.77\\
+CP+M+R       & 1.48           & 1.55          &  2.66         & 1.29         & 1.93 & 2.13 & 2.89 & 1.63 \\
+CP+M+R+CL    & \textbf{1.31}  & \textbf{1.37} &  \textbf{1.27}& \textbf{1.18}& \textbf{1.55} & \textbf{1.84} & \textbf{1.57} & \textbf{1.52} \\
\bottomrule
\end{tabularx}
\caption{Human evaluations on with average ranking of next step correctness (N.), future steps correctness (F.), diversity (D.), executability (E.). Ties are allowed.
\label{tab:human}
}
\end{table}

Since script learning is an open-ended task that is inherently difficult for automatic metrics to measure the correctness of generated scripts~\cite{huang2022language}, we further conduct a human evaluation. 
We hire four proficient English speakers as human annotators to independently rank the generation results from 1 (best) to 5 (worst) for: (1) \textit{next step correctness} which measures whether the generated results match the next step; (2) \textit{future steps correctness} measuring whether the generated results match any of the future steps; (3) \textit{diversity} which measures the diversity of generated results under the same subgoal; (4) \textit{executability} which checks the generated results repeat or conflict with step history. We randomly select ten subgoals, including 41 and 44 generated steps from the test set for Gardening and Crafts separately.

The human evaluation results\footnote{The Krippendorff-$\alpha$ inter-annotator agreement scores \citep{krippendorff2018content} and detailed guidelines of human evaluations are in the Appendix \ref{sec:humanevald}} are shown in Table \ref{tab:human}. Our contrastive learning model performs best over all metrics on two datasets. By adding each component of our model, we observe a consistent trend in correctness to ground truth. However, we also observe that scores for selective encoding decrease because the output space with selective encoding is more constrained than the BART baseline, and the length of our generated sequence is not very long.

\subsection{Discussions}

\noindent\textbf{Impact of Selective Multimedia Encoder} The caption input helps the model understand the general step descriptions better. 
For example, given the activity \textit{``cure azaleas of leaf gall''}, the step text only shows a generic instruction: \textit{``rule out other diseases''}. However, the BLIP captioner generates \textit{``a green leaf with white dots on it''} which helps the model generate \textit{``remove the leaf gall from the shrub''} instead of \textit{``keep your shrub healthy''}. 
Furthermore, in Figure \ref{img:task_example}, the finger object is absent from caption 3, indicating that the caption model has the ability to eliminate extraneous information from the image.
The selective gate can filter out unrelated steps which are not directly related to the current subgoal. 
For example, in Figure \ref{img:task_example}, our model successfully predicts a low masking weight of 0.049324 for the step “cut the thread”, while assigning a much higher masking weight of 0.134498 to its uninformative caption “a pair of scissors and a measuring tape”. The results imply that the selective gate successfully guides the model to focus on the related information.

\noindent\textbf{Impact of Retrieval Augmentation} The retrieved steps provide relevant knowledge from similar tasks: given the subgoal \textit{``finding or growing roses''} because the retrieved sentence mentioned \textit{``fertilizer''} and \textit{``mulch''}, the model successfully generates \textit{``fertilize your roses''}. Additionally, the model also benefits from retrieval augmentation with an analogy, e.g., the model generates \textit{``know when to harvest''} given the retrieved step \textit{``plant the bulbs when you get them''}. 

\noindent\textbf{Impact of Contrastive Learning} In addition to the improvement in diversity from the previous section, we observe that contrastive learning helps the model generate results closer to ground truth compared to other baselines. For example, it generates \textit{``pick creeping charlie plants from the ground''}, similar to ground truth \textit{``pick your creeping charlie leaves''}. The addition of contrastive learning also helps our model generates instructions with more details than other baselines by stating  \textit{``place the plant in the hole and cover it with soil''} instead of \textit{``place the plant in the hole''}.

\section{Related Work} 
Previous script learning tasks fall into two forms: selective and generative. 
The selective script learning tasks focus on modeling the script interactions given a list of candidates, including multi-choice goal step inference/ordering \citep{zhou-etal-2019-learning-household,zhang-etal-2020-reasoning}, script retrieval \citep{lyu-etal-2021-goal,zhou-etal-2022-show}, action anticipation \citep{Damen2018EPICKITCHENS,Damen2021RESCALING}, procedure segmentation \cite{8578725,10.5555/3504035.3504965,ghoddoosian2022hierarchical}, multi-choice visual goal-step inference \citep{yang-etal-2021-visual}, multimedia procedure planning~\citep{Zhao_2022_CVPR},  multimedia step ordering \citep{NEURIPS2021_c6d4eb15,wu-etal-2022-understanding}, instructional video retrieval \citep{ier}, and step classification \citep{Lin_2022_CVPR}. Despite promising results, their performance heavily relies on the given candidates, making them difficult to generalize for unseen activities. The second category is text-based generative script learning \citep{tandon-etal-2020-dataset,lyu-etal-2021-goal,huang2022language,li2020connecting,li2021future,jin2022event,sancheti-rudinger-2022-large}. However, this is the first work to provide a multimedia goal-oriented generative script learning along with a new multimodal-retrieval based metric. Different from \citet{Sener_2019_ICCV}, which uses a video to generate the next step, our new task uses step image-text pairs as input. Unlike previous multimedia script learning frameworks with a multimedia encoder to capture visual and textual information, we use a captioner to convert images into captions summarizing the important objects in images. 
The GOSC dataset~\cite{lyu-etal-2021-goal} contains the steps of daily stereotypical tasks, but most of the steps (52.6\%) in this dataset are unordered, making it infeasible to evaluate the next-step prediction. Consequently, we adapted the best model mT5 \cite{xue-etal-2021-mt5} in GOSC to our settings, i.e., the monolingual version T5, and used it as an equivalent baseline to show the comparison with the state-of-the-art model. 

To handle irrelevant sentences in the input, instead of using a token-level gating mechanism that only depends on the token itself \citep{sengupta-etal-2021-gated-transformer}, we introduce a sentence (step/caption) level gating mechanism whose gates depend on global context and weighted sentence representations. 
Our work is also related to retrieval-augmented text generation models~\citep{wang-etal-2019-paperrobot,NEURIPS2020_6b493230,liu-etal-2021-three}. However, instead of retrieving knowledge from an external corpus, we use steps from similar tasks in training data to guide the generation process. Moreover, we introduce a new contrastive learning loss to increase diversity. 
Previous contrastive learning-based text generation methods usually use negative samples constructed by sequence manipulation \citep{cao-wang-2021-cliff,hu-etal-2022-planet} or perturbation \citep{lee2021contrastive}. Inspired by \citet{wang-etal-2022-simkgc} which uses self-negatives for knowledge graph completion and that the generation output tends to repeat the input, we extend self-negatives for sequence-to-sequence contrastive learning. We also retrieve similar steps from the training set as additional hard negatives.

\section{Conclusion}
We propose a novel Multimedia Generative Script Learning task with the first benchmark featuring step and descriptive image pairs to generate subsequent steps given historical states in both text and vision modalities. Moreover, we build a new script learning framework consisting of a selective multimedia encoder, a retrieval-augmented decoder, and a diversity-oriented contrastive learning objective to generate the next steps. Furthermore, we define a new \textit{multimodal-retrieval based metric} which can be used for multimedia script learning tasks. Automatic and human evaluation results demonstrate consistent performance improvements.

\section{Limitations}
\subsection{Limitations of Data Collection}
Regarding data collection, we crawled the English WikiHow website from Jan 2021 to May 2021. The number of available activities is limited by the data we crawled from WikiHow. We currently only choose \textit{Gardening} and Crafts categories as case studies. Because we focus on multimedia image-step pairs, we remove steps \textit{that} are not attached to any illustrative images. We also observe that a small portion of activities in the dataset do not follow chronological order.

Since our task focuses on the daily stereotypical tasks which usually require the model to understand the visual environment, the model design can be directly applied to support other domains, such as steps in the cooking videos. In addition, our model can also adapt to scenarios without visual images because the performance of our model only decreases slightly if no caption is provided. 
We plan to expand our model to other categories written in other languages. 

\subsection{Limitations of System Performance}
The model might generate incorrect nouns because  of  the occurrence of patterns (e.g., \textit{``refrigerate the \textbf{slane} for up to 1 year''} instead of \textit{``refrigerate the \textbf{purslane} for up to 1 year''}).  In addition, our model sometimes tends to generate generic step descriptions because of insufficient input information, e.g., given the last step \textit{``lay the t-shirt out on a clean, flat surface.''}, the model generates \textit{``cut the shirt out''} which is vague compared to ground truth \textit{``carefully cut around the sleeve''}. Moreover, the pretrained model might focus more on language modeling instead of inherent logic: for the activity of \textit{``make paint can planters''}, after \textit{``removing the label''} from the paint can, the BART+CAP generates \textit{``read the label''}. In addition, there is still a small chance that the model generates the same output for various similar inputs.

Because we rely on image captions and retrieval results for step prediction, the upper bound of our generation quality is limited by the performance of the image caption and sentence retrieval modules. Our framework also needs to improve on imbalanced topics in the dataset. For example, the dataset contains more activities about \textit{tree} for the gardening domain than other gardening-related plants. Because our multimedia generative script learning is a new task, we cannot compare our model with other established state-of-the-art models. Moreover, because WikiHow is a crowd-sourcing website, some everyday activities might have better human annotations than the remaining activities. We plan to include a fine-grained human written step prediction as an upper bound to address this issue.

\subsection{Limitations of Evaluation} 

The automatic metrics we chose, including BLEU \cite{papineni-etal-2002-bleu}, ROUGE \cite{lin-2004-rouge}, METEOR \cite{denkowski-lavie-2014-meteor}, BARTScore \cite{NEURIPS2021_e4d2b6e6}, self-BLEU \cite{10.1145/3209978.3210080}, and unique $n$-grams \cite{fedus2018maskgan}, might not be the best metrics to evaluate our results. Some other metrics, such as semantic similarity and multimodal-retrieval based metrics, are based on pretrained models, including Augmented SBERT \cite{thakur-etal-2021-augmented}, SentenceBert \cite{reimers-gurevych-2019-sentence}, and CLIP \cite{pmlr-v139-radford21a}. Those metrics might not align with human judgment and might be biased toward pretrained datasets. While we complement it with human evaluation, we only focus on relevance to ground truth and diversity. Although we found fluency is not an issue, it is likely we still need to cover all aspects of generation results.

\section{Ethics and Broader Impact}
The type of multimedia script learning framework we have designed in this paper is limited to WikiHow articles, and they might not be applicable to other scenarios. 
\subsection{Usage Requirement}
Our multimedia script learning framework provides investigative leads for multimedia script prediction. Therefore, it is not intended to be used for any activity related to any human subjects. Instead, our system aims to generate step predictions with unseen activities similar to those in the training set. Accordingly, domain experts might use this tool as an assistant to write more constructive instructional scripts that would be too time-consuming for a human to create from scratch. 
Experts can also use this system to improve writing instruction by adding missing instructions. However, our system does not perform fact-checking or incorporate any external knowledge, which we leave as future work. The IRB board should first approve human subjects who follow instructions generated by our system. 

\subsection{Data Collection}
We collect data by crawling the raw official English WikiHow website, which is under \textit{Attribution-Noncommercial-Share Alike 3.0 Creative Commons License}\footnote{\url{https://www.wikihow.com/wikiHow:Creative-Commons}}. We ensure that our data collection procedure follows the Terms of Use located at \url{https://www.wikihow.com/wikiHow:Terms-of-Use}. Therefore our dataset can only be used for non-commercial purposes. As mentioned in Section \ref{sec:human}, we perform the human evaluation. All annotators involved in the human evaluation are voluntary participants and receive a fair wage.

\section*{Acknowledgement}
This work is supported by Agriculture and Food Research Initiative (AFRI) grant no. 2020-67021-32799/project accession no.1024178 from the USDA National Institute of Food and Agriculture, and by U.S. DARPA KAIROS Program No. FA8750-19-2-
1004. The views and conclusions contained herein are those of the authors and should not be interpreted as necessarily representing the official policies, either expressed or implied of the U.S. Government. The U.S. Government is authorized to reproduce and distribute reprints for governmental purposes notwithstanding any copyright annotation therein.
Hou Pong Chan was supported in part by the Science and Technology Development Fund, Macau SAR (Grant Nos. FDCT/060/2022/AFJ, FDCT/0070/2022/AMJ) and the Multi-year Research Grant from the University of Macau (Grant No. MYRG2020-00054-FST). 
\bibliography{anthology,custom}
\bibliographystyle{acl_natbib}

\appendix
\section{Hyperparameters}
\label{sec:hyper}
 Our model is built based on the Huggingface framework \citep{wolf-etal-2020-transformers}\footnote{https://github.com/huggingface/transformers}. We choose top $5$ retrieved historically relevant steps as input for our retrieval model. We choose $5$ negative samples for each step during contrastive learning for the gardening domain. Specifically, $4$ self-negative samples, including steps and captions, are randomly chosen from the title, method, and step history input. The remaining $1$ retrieved negative samples are randomly chosen from top-$20$ most similar steps retrieved from the training set based on the last step. For the crafts domain, we choose $5$ self-negative samples and $5$ retrieved negative samples. We set $\tau$ as 1 for contrastive loss and $\lambda$ as 0.5 based on validation performance for the training objectives. We optimize our model by AdamW \citep{Loshchilov2019DecoupledWD} with Cosine Annealing Warm Restarts schedule \citep{DBLP:conf/iclr/LoshchilovH17}. Our learning rate is $1\times 10 ^{-5}$ with  $\epsilon=1\times 10 ^{-6}$ for gardening domain and $2\times 10 ^{-5}$ with  $\epsilon=1\times 10 ^{-6}$ for crafts domain. The number of warm-up steps is $2000$. The batch size is set to $16$ for both domains, and the maximum training epoch is set as $30$ with $10$ patience. During decoding, we use beam-search to generate results with a beam size of 5.
 
\section{Training details}
\label{sec:train}

\begin{table}[!htb]
\centering
\small
\begin{tabularx}{\linewidth}{>{\hsize=1\hsize}X>{\centering\arraybackslash\hsize=1\hsize}X}
\toprule
&\textbf{\# of Parameters}  \\
\midrule
BART                           & 139.425M \\ 
\hspace{1mm}+CP               & 139.425M\\ 
\hspace{1mm}+CP+M            & 141.788M\\ 
\hspace{1mm}+CP+M+R         & 158.346M\\ 
\hspace{1mm}+CP+M+R+CL      & 158.347M \\
\bottomrule
\end{tabularx}
\caption{\# of Model Parameters \label{tab:para}}
\end{table}

 We use BART-base from Huggingface \cite{wolf-etal-2020-transformers} for our method and baselines. We normalize all our input sentences into lower case.
 We add 5 special tokens for BART-base model including $<$title$>$, $<$method$>$, $<$step$>$, $<$caption$>$, $<$template$>$, and $<$cls$>$. We prepend $<$title$>$ to goal, $<$method$>$ to subgoal, $<$step$>$ to text step, $<$caption$>$ to step caption, $<$template$>$ to retrieved step, and <cls> to the beginning of step history input. We truncate our step, caption, goal, and subgoal to 30 tokens and target step to 40. We only choose the closest 10 step-caption pairs. We use BLIP~\citep{li2022blip} \footnote{The BLIP checkpoint we is \url{https://storage.googleapis.com/sfr-vision-language-research/BLIP/models/model_base_capfilt_large.pth}} pretrained with 129M images including including COCO \citep{10.1007/978-3-319-10602-1_48}, Visual Genome \citep{10.1007/s11263-016-0981-7}, Conceptual Captions \citep{sharma-etal-2018-conceptual}, Conceptual 12M \citep{Changpinyo_2021_CVPR}, SBU \citep{NIPS2011_5dd9db5e}, and LAION \citep{400m}. We use $\mathtt{all-mpnet-base-v2}$ from SentenceBert \citep{reimers-gurevych-2019-sentence}, which performs best in semantic search to retrieve similar steps. 
 
 \begin{table}[!htb]
\centering
\small
\begin{tabularx}{\linewidth}{>{\hsize=1\hsize}X>{\centering\arraybackslash\hsize=1\hsize}X>{\centering\arraybackslash\hsize=1\hsize}X>{\centering\arraybackslash\hsize=1\hsize}X}
\toprule
\textbf{\# History } & \textbf{\# Instance} &\textbf{BARTScore}$\uparrow$&\textbf{Semantic}$\uparrow$\\
\midrule
1&685 & -4.3683 & 0.3189  \\
2&680 & -4.3633 & 0.3115 \\
3&545 & -4.4213 & 0.3064 \\
4&346 & -4.3535 & 0.3118 \\
5&207 & -4.3556 & 0.2748 \\
6&104 & -4.3588 & 0.2746 \\
7&56 & -4.2192 & 0.3381 \\
8&26 & \textbf{-4.1687} & \textbf{0.3411} \\
9&12 & -4.3800 & 0.2085 \\
10&23 &-4.7718 & 0.2491 \\
\bottomrule
\end{tabularx}
\caption{The average number of BARTScore/ Semantic Similarity Score and the number of instances given the different lengths of step history for the gardening domain\label{tab:nstp}}
\end{table}
 We train our model with NVIDIA A6000 GPUs with 48G memory with full precision. We choose our best model based on the validation score with BLEU-4 \citep{papineni-etal-2002-bleu} and ROUGE \citep{lin-2004-rouge}. The best validation scores for our contrastive learning model are: BLEU-4 with $2.81$ and ROUGE-L with $15.24$ for the gardening domain; BLEU-4 with $4.85$ and ROUGE-L with $20.25$ for the crafts domain. The average training time for each model is 2 to 4 hours. Table \ref{tab:para} shows the number of parameters for each model.
 
\begin{table*}[!htb]
\centering
\small
\begin{tabularx}{\linewidth}{>{\hsize=1\hsize}X>{\hsize=1.6\hsize}X>{\centering\arraybackslash\hsize=0.7\hsize}X>{\centering\arraybackslash\hsize=0.7\hsize}X>{\centering\arraybackslash\hsize=0.7\hsize}X>{\centering\arraybackslash\hsize=0.7\hsize}X>{\centering\arraybackslash\hsize=1\hsize}X>{\centering\arraybackslash\hsize=0.8\hsize}X>{\centering\arraybackslash\hsize=1.4\hsize}X>{\centering\arraybackslash\hsize=1.4\hsize}X}
\toprule
\textbf{Domain}&\textbf{Model}&\textbf{B-1}$\uparrow$&\textbf{B-2}$\uparrow$&\textbf{B-3}$\uparrow$&\textbf{B-4}$\uparrow$&\textbf{METEOR}$\uparrow$&\textbf{R-L}$\uparrow$&\textbf{BARTScore}$\uparrow$&\textbf{Semantic}$\uparrow$\\ 
\midrule
\multirow{3}{*}{Gardening}
&BART+CP+M+CL&\textbf{17.9}  &\textbf{9.30}  &\textbf{5.20}  &\textbf{3.07}  &\textbf{9.72} &\textbf{17.1}  &\textbf{-4.39} &0.304 \\
&BART+CP+R+CL&17.6           &9.16           &5.16           &3.03  &9.54 &16.7  &-4.41 &0.299 \\
&BART+M+R+CL &17.7           &9.11           &4.98           &2.92  &9.71 &17.0  &-4.37 &\textbf{0.306}\\
\hdashline
\multirow{3}{*}{Crafts}
&BART+CP+M+CL&20.6  &10.9  &6.12  &3.89  &10.8 &19.3  &-4.30 &\textbf{0.307}\\
&BART+CP+R+CL&20.3  &11.0  &6.36  &4.12  &10.8 &19.8   &-4.29 &0.301\\
&BART+M+R+CL &\textbf{20.8}  &\textbf{11.5}  &\textbf{6.78}  &\textbf{4.49}  &\textbf{10.9} &\textbf{20.1}  &\textbf{-4.27} &0.306\\

\bottomrule
\end{tabularx}
\caption{Automatic evaluation results on next step prediction for the gardening and crafts domain (\%).
\label{tab:step_a} }
\end{table*}

\section{Evaluation Metrics}
\label{sec:evalmetrics}
 We use BLEU \citep{papineni-etal-2002-bleu}, ROUGE \citep{lin-2004-rouge}, and METEOR \citep{denkowski-lavie-2014-meteor} from  Microsoft COCO Caption Evaluation package\footnote{\url{https://github.com/salaniz/pycocoevalcap}}. We use official implementation of BARTScore \citep{NEURIPS2021_e4d2b6e6}\footnote{\url{https://github.com/neulab/BARTScore}}. We use $\mathtt{cross-encoder/stsb-roberta-large}$ which performs best on STSBenchmark \citep{cer-etal-2017-semeval} to compute semantic similarity score from Augmented SBERT \citep{thakur-etal-2021-augmented}. For multimodal-retrieval based metric, we use the best sentence embedding model: $\mathtt{all-mpnet-base-v2}$ from SentenceBert \citep{reimers-gurevych-2019-sentence} for text retrieval, and the best language-image pretraining model $\mathtt{ViT-L/14@336px}$ from CLIP \citep{pmlr-v139-radford21a} for image retrieval. 
 Specifically, we compute the CLIP similarity between the image embedding and the sentence embedding of the target step to retrieve images.
 All results are based on a single run. 
 We have opted not to include a human performance baseline in our evaluation. This decision was made due to the inherent challenges of assessing human performance in generative script learning, which requires annotators to possess domain knowledge in order to predict the next steps accurately. Moreover, different tasks may require different levels of expertise, experience, or background knowledge, making it difficult to establish a consistent baseline for human performance evaluation.
\section{Additional Ablation Study}
\label{sec:abl}

We conducted further ablation experiments, the results of which are presented in Table \ref{tab:step_a}. Our findings show that all ablated models performed worse than our proposed model.

\section{Prediction for different history length}

\begin{figure}[ht]

     \centering
     \begin{subfigure}[b]{0.49\textwidth} 
     \includegraphics[width=\textwidth]{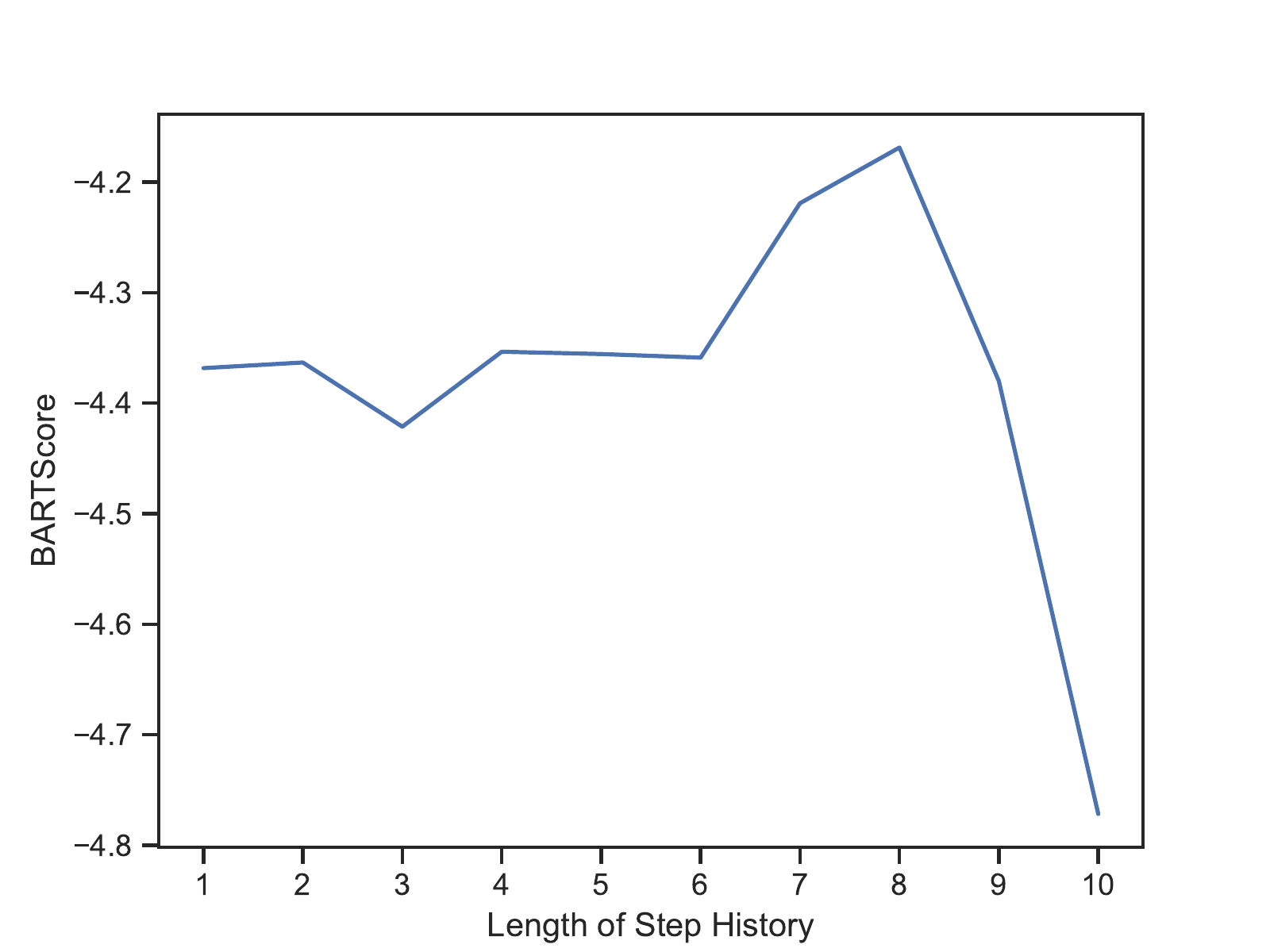}
     \caption{The average number of BARTScore in test set given the different lengths of step history}
      \label{img:bart}
     \end{subfigure}
     \hfill
     \begin{subfigure}[b]{0.49\textwidth}
     \includegraphics[width=\textwidth]{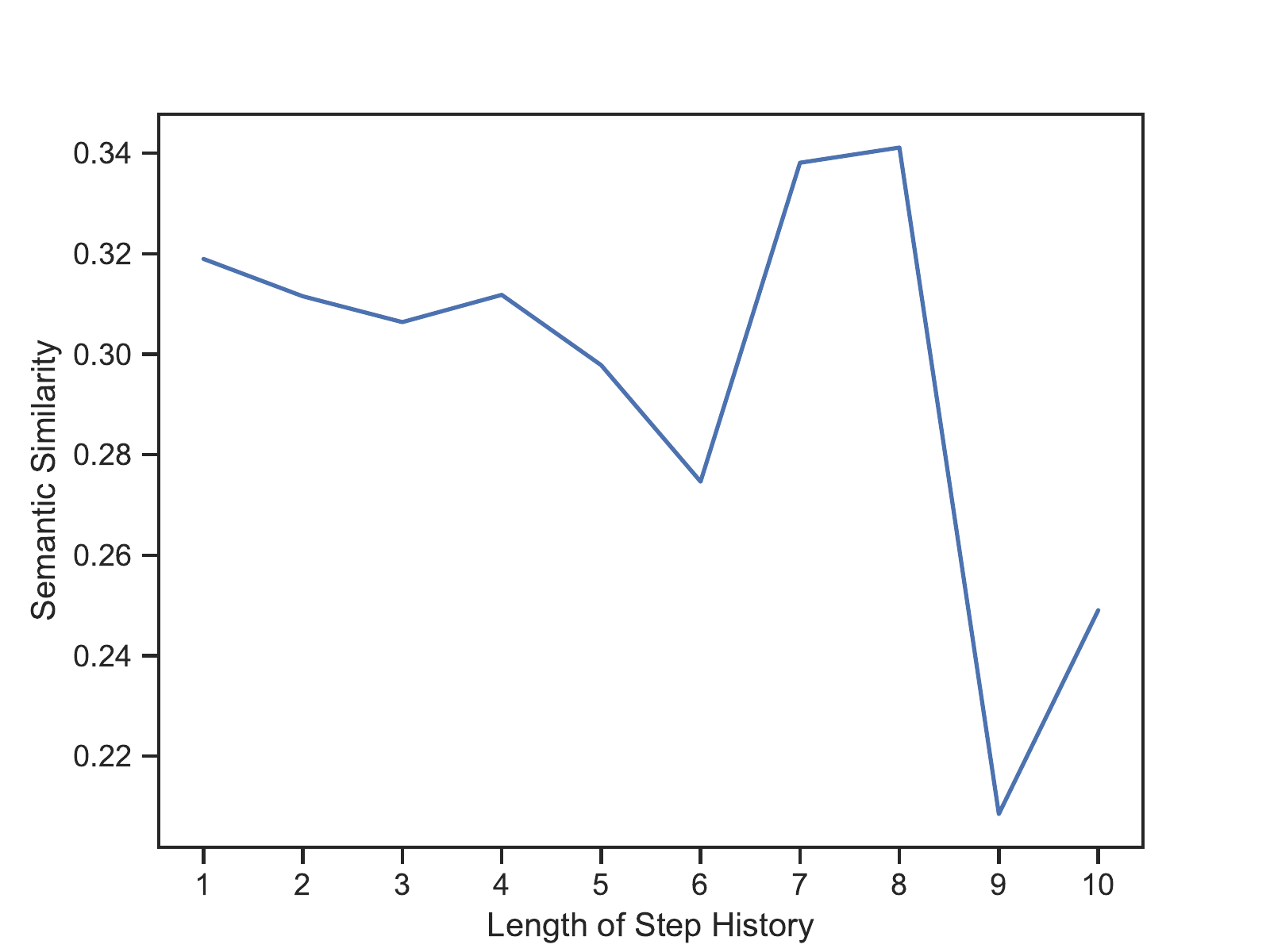}
     \caption{The average number of Semantic Similarity Score in test set given the different lengths of step history}
      \label{img:sim}
     \end{subfigure}

        \caption{Prediction for different history lengths for the gardening domain}
\end{figure}

In Figure \ref{img:bart} and Figure \ref{img:sim}, we show the averaged BARTScore and semantic similarity scores of our contrastive learning models in the next step prediction task over different step history lengths.  
In both figures, we observe that the results with eight step-caption pairs obtain the highest scores. We analyze the reasons as follows. For the instances that contain less than eight history steps, 
increasing the step history introduces more information than noise from the step text and corresponding captions. 
However, as the step length grows, the additional step-caption pairs introduce more noise than information relevant to the future step. 
Empirically, the eight-step length achieves an optimal balance between noise and relevant information. Another potential reason is related to the number of instances. In Table \ref{tab:nstp}, we see a clear decline in the number of instances because of our dataset construction strategy. Therefore, the model cannot generalize over long history input.

\section{Dataset Collections}
We crawled the English WikiHow website from Jan 2021 to May 2021.
We extract all articles from the crawled website dump in the \textit{Gardening} and \textit{Crafts} categories. Each article contains a unique activity. We use BeautifulSoup \citep{richardson2007beautiful} to parse the article and obtain JSON files. Each JSON file contains a gardening activity. For each gardening activity, we remove those steps without paired images or steps whose images do not exist in the dump. Then, we use a regular expression to remove the URLs in the steps. We remove those steps that are too short (less than two words) or contain no values. Finally, we remove the activity containing only one step in each subgoal. 

\section{Parallel Steps}
In this paper, we focus on predicting correct orders for sequential step prediction since we find that only 18\% of the subgoals have one parallel step by random checking 50 subgoals, and 14\% contain more than one parallel step. It is more critical to predict correct orders for non-interchangeable steps, such as step 4 and 5 in Figure \ref{img:task_example}.  By using generative methods, multiple steps can be predicted with different probabilities, which can support parallel processes. We also propose the multimodal-retrieval-based metric by treating the future steps as a set and checking whether the generation steps fall into the future steps.

\begin{figure}[htb]

     \centering
     \includegraphics[width=0.5\textwidth]{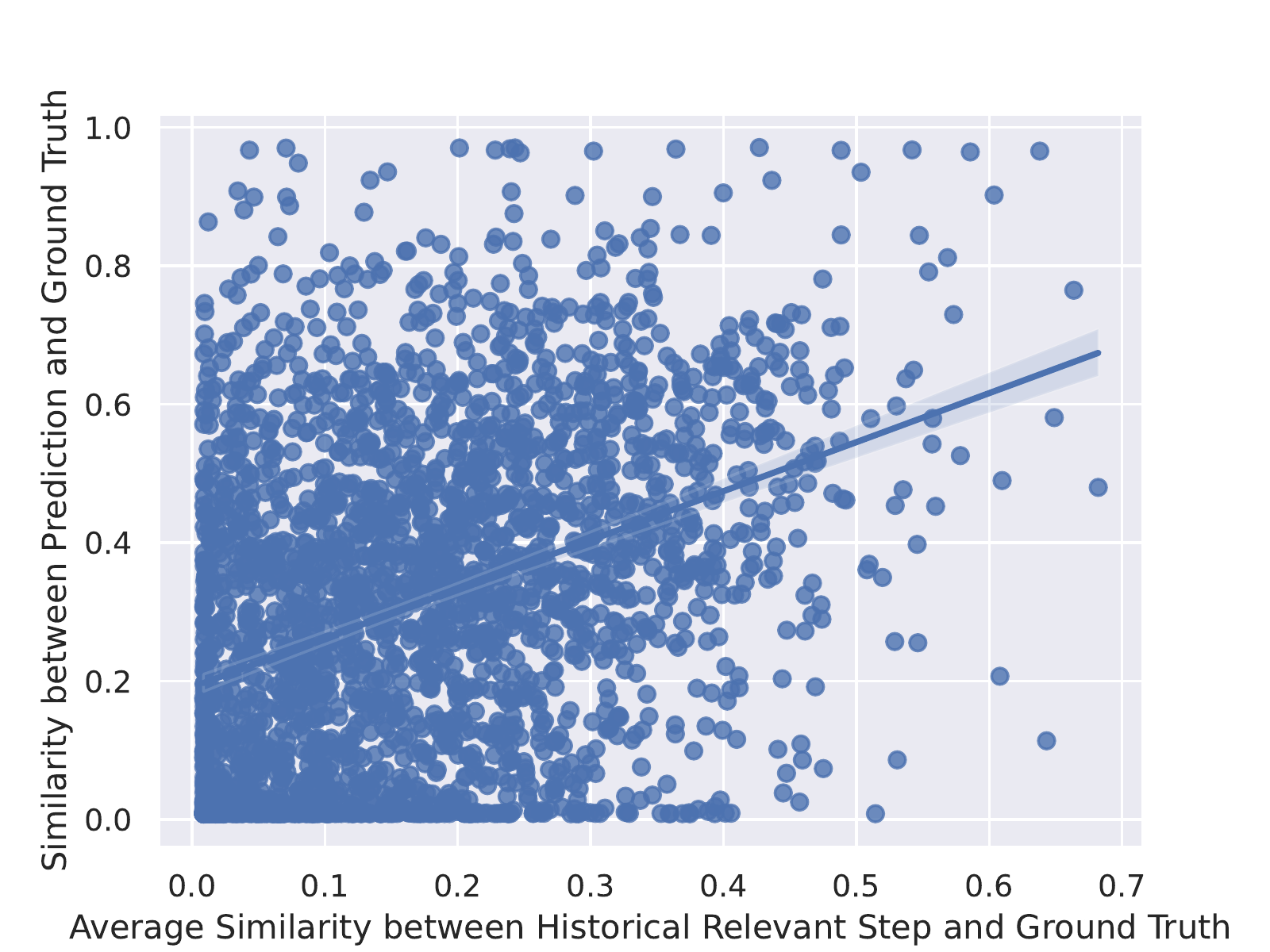}
     \caption{
     The semantic similarity scores \citep{thakur-etal-2021-augmented} between the model predictions and the ground truths versus the semantic similarity scores between the retrieved historical relevant steps and the ground truths in the gardening domain. 
     }
        \label{fig:hg_avg}
\end{figure}

\section{Impact of Historical Relevant Steps}

We analyze the relation between the quality of the retrieved historically relevant steps and the quality of the model predictions. 
The semantic similarity score evaluates the quality of retrieved steps and model predictions, which measures the embedding space similarity between a given text and the ground-truth next step. 
Pearson's correlation between the semantic scores of historically relevant steps and the semantic scores of model predictions is 0.39 with a $p<0.01$. We also illustrate their relation in Figure \ref{fig:hg_avg}. The results suggest that the performance of our model is positively correlated with the relevance of the retrieved historical steps.

\begin{figure}[htb]

     \centering
     \includegraphics[width=\linewidth]{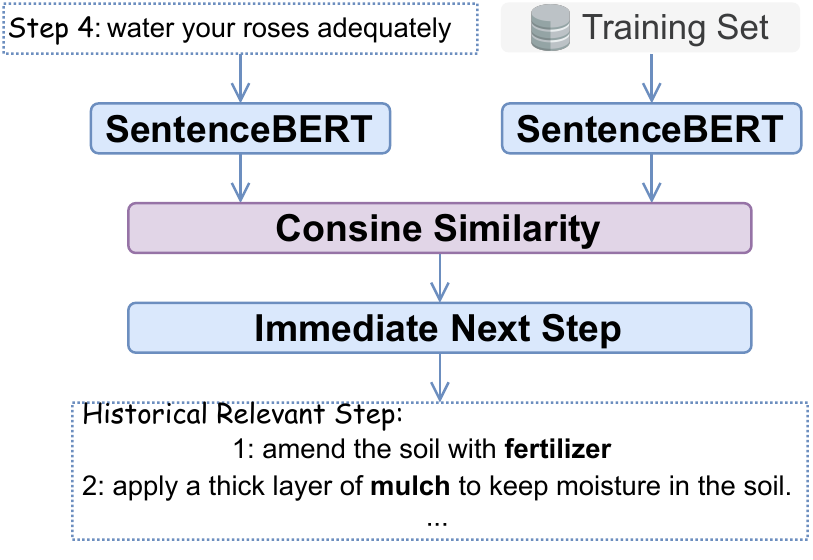}
     \caption{Details for historical relevant step retrieval}
      \label{img:ret_detal}
\end{figure}

\section{Additional Model Architecture}

\begin{figure}[htb]
     \includegraphics[width=\linewidth]{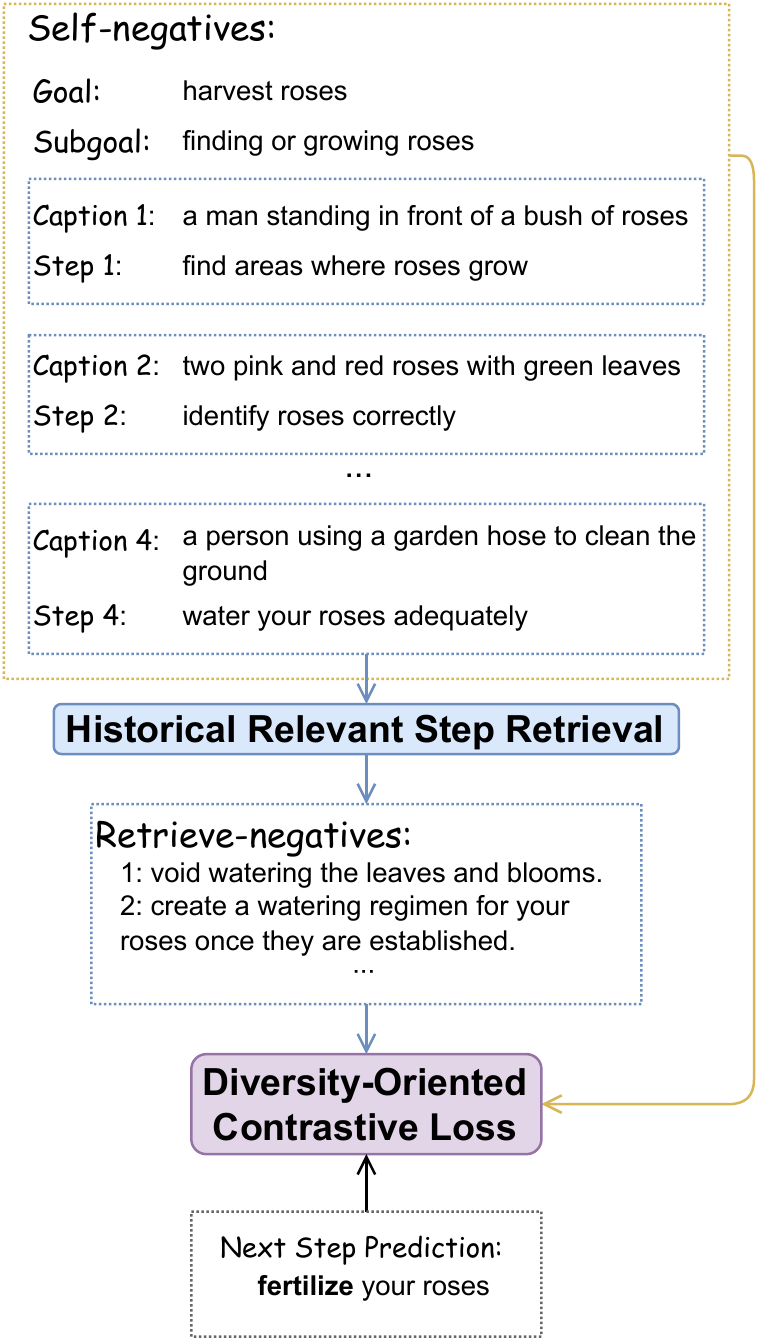}
     \caption{Details for diversity-oriented contrastive loss}
      \label{img:div_det}
     
\end{figure}
Figure \ref{img:ret_detal} and \ref{img:div_det} show additional details for our framework.  The immediate next step refers to the step right after the previously given steps. 

\section{Scientific Artifacts}
We list the licenses of the scientific artifacts used in this paper: WikiHow (Attribution-Noncommercial-Share Alike 3.0 Creative Commons License), Huggingface Transformers (Apache License 2.0), SBERT (Apache-2.0 license), BARTScore (Apache-2.0 license), CLIP (MIT license), and BLIP (BSD-3-Clause license).

\newpage
\section{Human evaluation details}
\label{sec:humanevald}
\begin{table}[!htb]
\centering
\small
\begin{tabularx}{\linewidth}{>{\hsize=4.2\hsize}X>{\centering\arraybackslash\hsize=0.6\hsize}X>{\centering\arraybackslash\hsize=0.6\hsize}X>{\centering\arraybackslash\hsize=0.6\hsize}X>{\centering\arraybackslash\hsize=0.6\hsize}X>{\centering\arraybackslash\hsize=0.6\hsize}X>{\centering\arraybackslash\hsize=0.6\hsize}X>{\centering\arraybackslash\hsize=0.6\hsize}X>{\centering\arraybackslash\hsize=0.6\hsize}X}
\toprule
\multirow{ 2}{*}{\textbf{Model}}&\multicolumn{4}{c}{\textbf{Gardening}}&\multicolumn{4}{c}{\textbf{Crafts}}\\
&\textbf{N.}&\textbf{F.}&\textbf{D.}&\textbf{E.}&\textbf{N.}&\textbf{F.}&\textbf{D.}&\textbf{E.}\\
\midrule
BART              & 0.60  & 0.64 &  0.55 & 0.22& 0.60 &0.59 &0.70 &0.35\\
+CP               & 0.65  & 0.50 &  0.53 & 0.41& 0.67 &0.60 &0.90 &0.31\\
+CP+M            & 0.70  & 0.74 &  0.86 & 0.31& 0.45 &0.40 &0.76 &0.41\\
+CP+M+R         & 0.53  & 0.50 &  0.68 & 0.37& 0.62 &0.46 &0.78 &0.31\\
+CP+M+R+CL      & 0.43  & 0.58 &  0.56 & 0.26& 0.58 &0.48 &0.13 &0.35 \\
\bottomrule
\end{tabularx}
\caption{Krippendorff-$\alpha$ scores for human evaluation on with average ranking of next step correctness (N.), future steps correctness (F.), diversity (D.), executability (E.).  \label{tab:alpha} }
\end{table}

 We measure inter-annotator agreement with Krippendorff-$\alpha$ scores \citep{krippendorff2018content}. The results are in Table \ref{tab:alpha}. Table \ref{tab:annotation} shows the annotation examples. Because we do not have a virtual environment to execute those steps, we do not have a good inter-annotator agreement on the executability.  
 
 \begin{table*}[!htb]
\small
\begin{tabularx}{\linewidth}{>{\hsize=0.15\hsize}X>{\arraybackslash\hsize=0.85\hsize}X}
\toprule
\textbf{Type}&\textbf{Content}\\
\midrule
Instructions& (1) similarity to the next step measures the correctness of generated results with the next step; (2) similarity to future steps measures whether the generated results are relevant to the future steps; (3) diversity measures the diversity of generated results under the same subgoal (4) executability which checks the generated results repeat or conflict with step history/
Please rank these models' output from 1(best)-5(worst), ties are allowed if both outputs are the same. \\\hdashline
Similarity and executability annotation examples & 

\textit{Title:}\newline
    protect garden berries\newline
\textit{Subgoal:}\newline
	setting up decoys\newline
\textit{Step History:}\newline
	use plastic snakes.\newline
	----------------------------------------\newline
\textit{Ground Turth Target:}\newline
\textit{Next Step:}\newline
	put out shiny pinwheels.\newline
\textit{Future Steps:}\newline
	put out shiny pinwheels.\newline
	create a decoy food area.\newline
	----------------------------------------\newline
\textit{Predictions:}\newline
\textit{0's prediction:}\newline
	wrap the snake in a plastic bag.\newline
\textit{1's prediction:}\newline
	set up a trellis.\newline
\textit{2's prediction:}\newline
	cut the berries down to the ground.\newline
\textit{3's prediction:}\newline
	set up a trap.\newline
\textit{4's prediction:}\newline
	choose a sturdy piece of string.\newline
\\\hdashline
Diversity&
\textit{0's predictions:}\newline
	wrap the snake in a plastic bag.\newline
	place the flowers on a stick in the dirt.\newline
\textit{1's predictions:}\newline
	choose the right plant.\newline
	set up a trap.\newline
\textit{2's predictions:}\newline
	cut the berries down to the ground.\newline
	create a trap.\newline
\textit{3's predictions:}\newline
	set up a trap.\newline
	create a trap.\newline
\textit{4's predictions:}\newline
	choose a sturdy piece of string.\newline
	set up a trap.\newline
\\
\bottomrule
\end{tabularx}
\caption{Annotation examples \label{tab:annotation}}
\end{table*}

\newpage
\section{Sample Output}
 \begin{figure}[!hbtp]
\centering
\includegraphics[width=\linewidth]{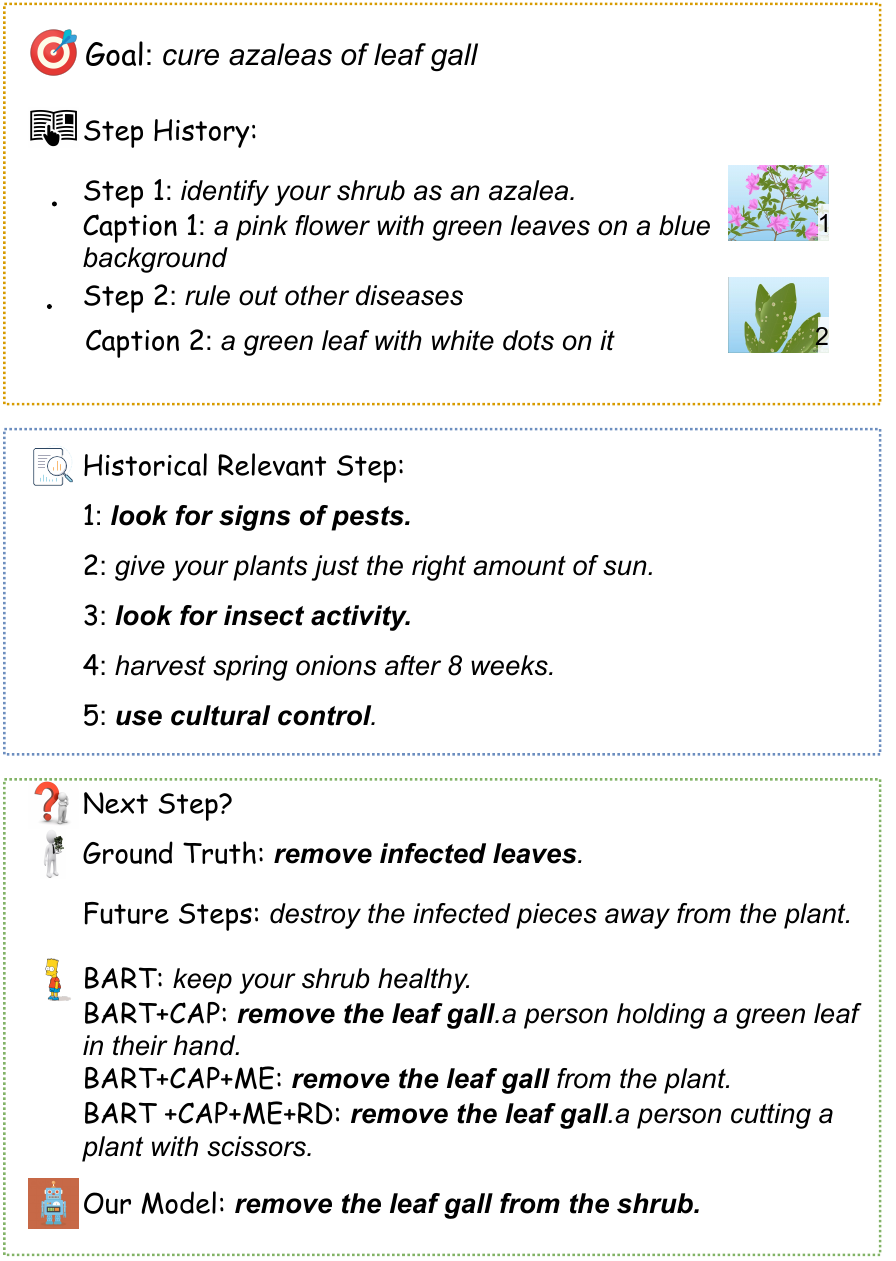}
\caption{Human and System Step Prediction Results. It shows an example that our model benefits from selective multimedia encoder.} 
\label{img:3}
\end{figure}

 \begin{figure}[!hbtp]
\centering
\includegraphics[width=\linewidth]{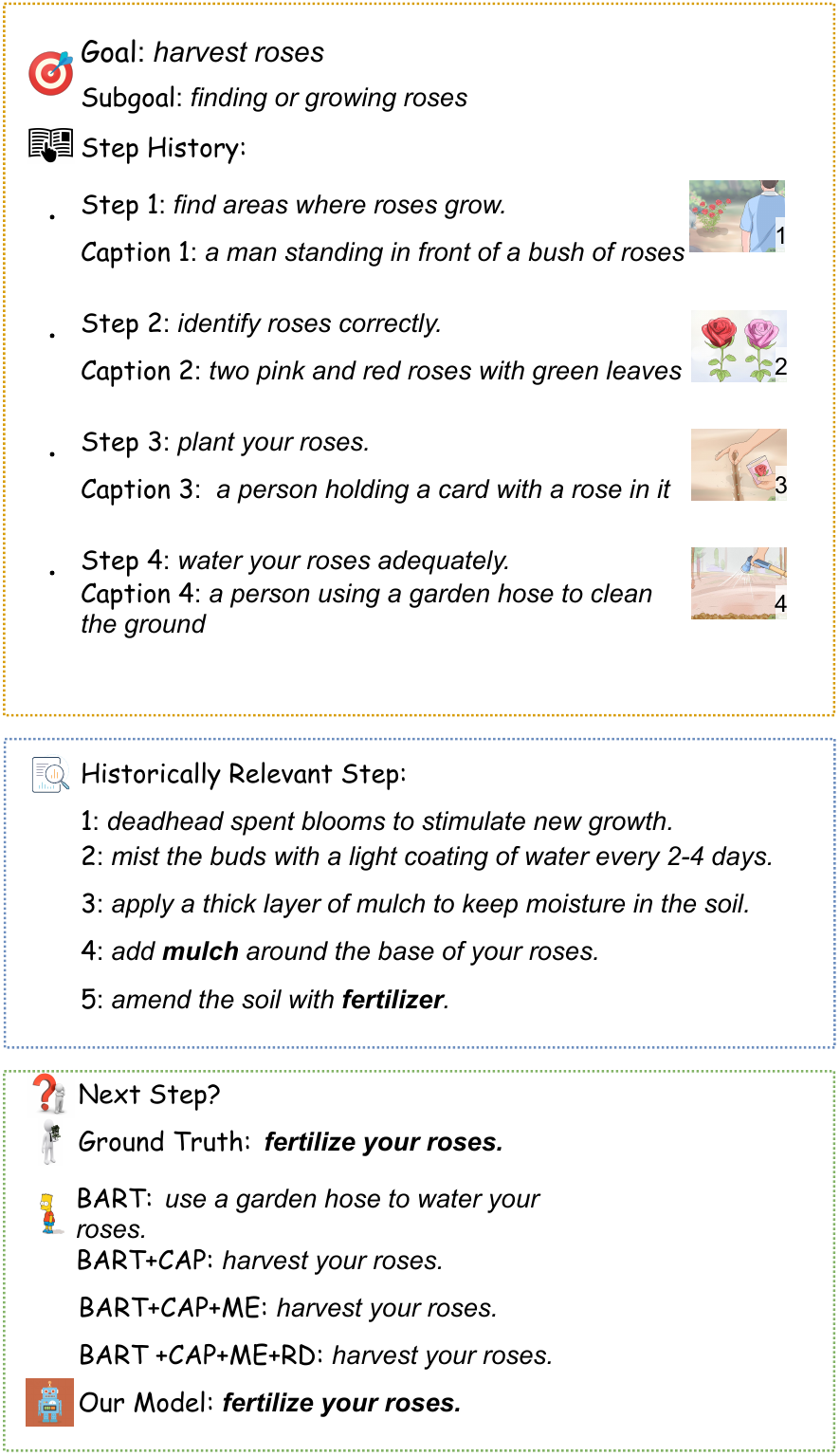}
\caption{Human and System Step Prediction Results. It shows an example that our model prediction results benefits from retrieval results and contrastive learning.} 
\label{img:4}
\end{figure}

 \begin{figure}[!hbtp]
\centering
\includegraphics[width=\linewidth]{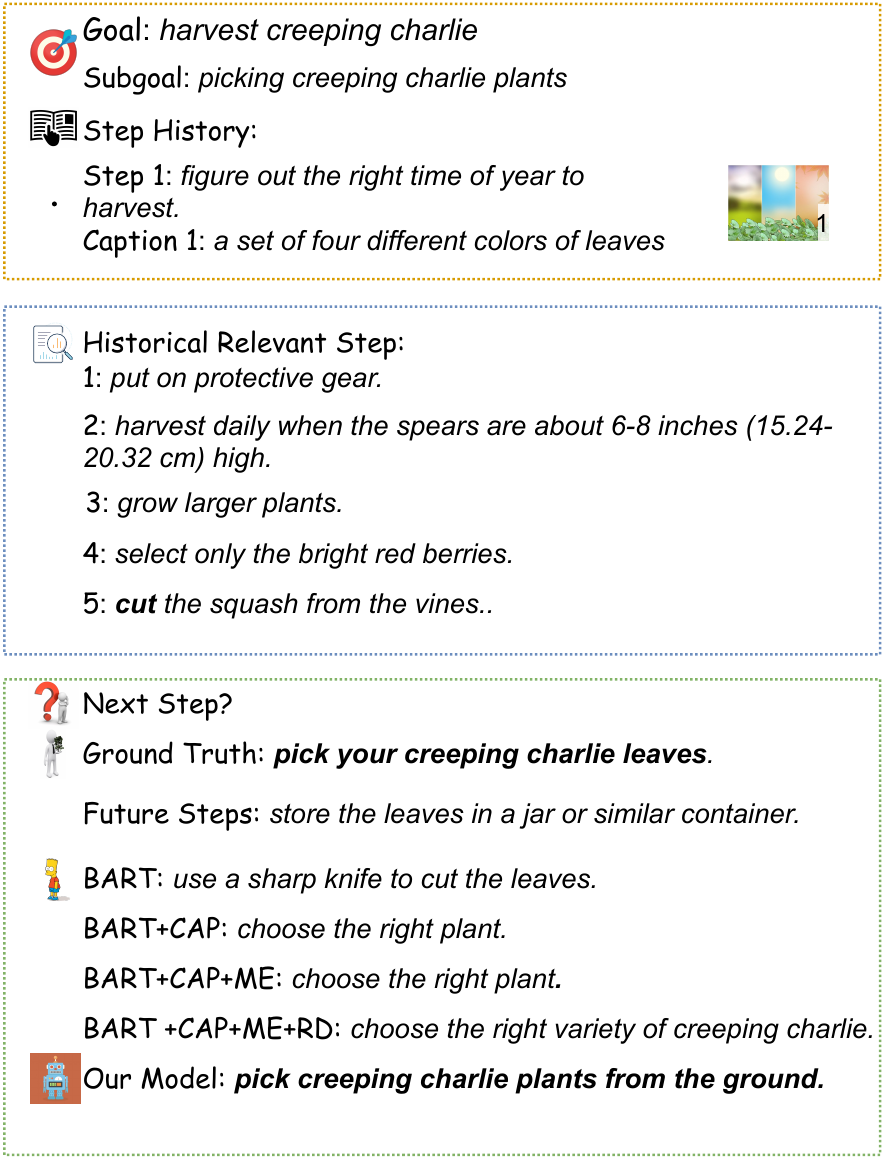}
\caption{Human and System Step Prediction Results. It shows an example that our model prediction results benefits from retrieval results and contrastive learning.} 
\label{img:2}
\end{figure}

 \begin{figure}[!hbtp]
\centering
\includegraphics[width=\linewidth]{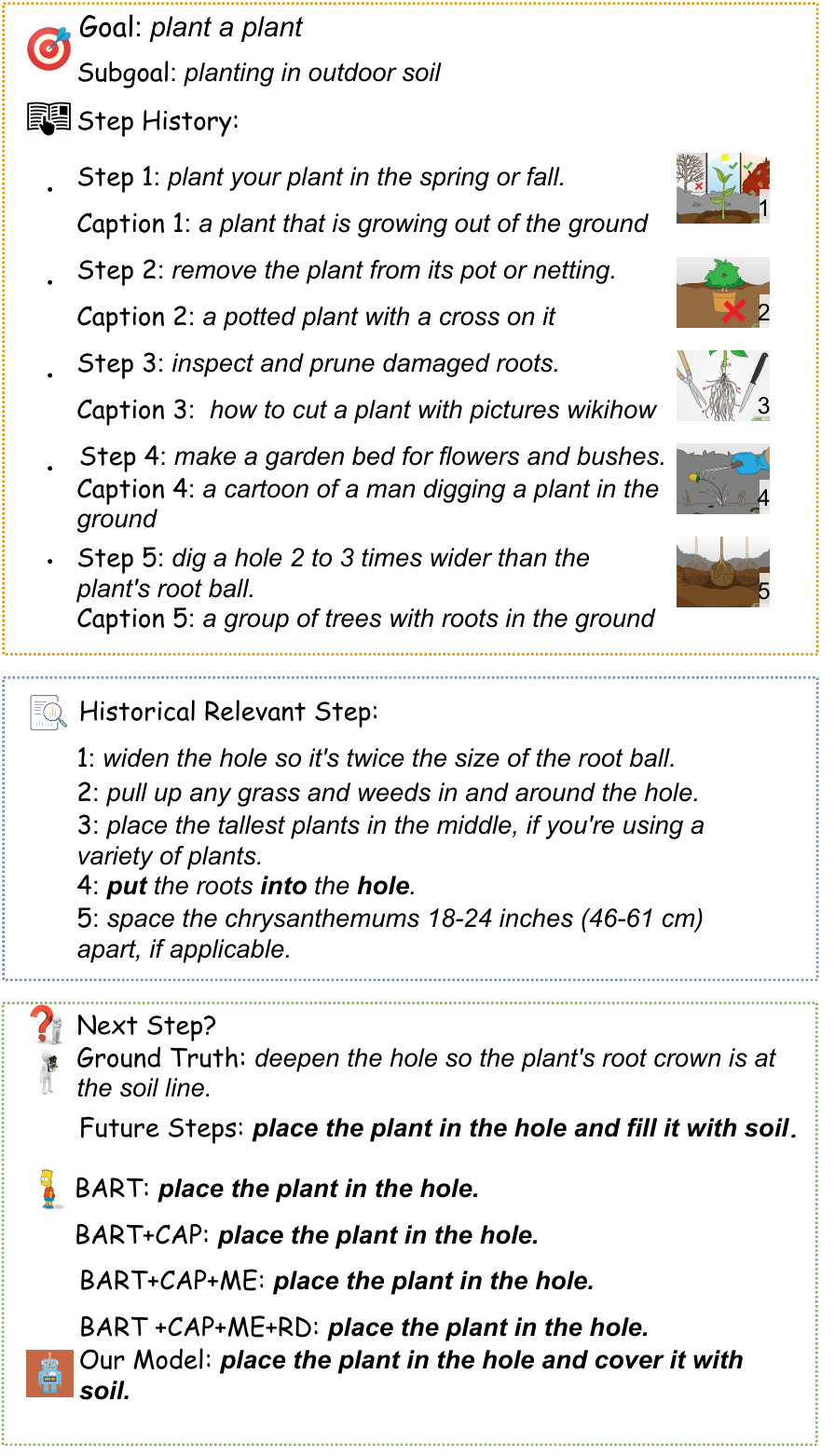}
\caption{Human and System Step Prediction Results. It shows an example that our model prediction results matches future steps instead of immediate next step.} 
\label{img:1}
\end{figure}

\end{document}